\documentclass[11pt]{article}
\usepackage{coling2020}
\usepackage{times}
\usepackage{url}
\usepackage{latexsym}
\usepackage{float}
\usepackage{graphicx}
\usepackage{tikz}
\usepackage{amsmath,amssymb}
\usepackage{hyperref}
\usepackage{caption}

\DeclareRobustCommand{\bigO}{%
  \text{\usefont{OMS}{cmsy}{m}{n}O}%
}
\DeclareMathOperator{\sign}{sign}

\colingfinalcopy 

\title{Unsupervised Sign Language Phoneme Clustering\\using HamNoSys Notation}

\author{Boris Mocialov \\
  School of Engineering \\ and Physical Sciences \\
  Heriot-Watt University \\
  {\tt bm4@hw.ac.uk} \\\And
  Graham Turner \\
  School of Social Sciences \\
  Heriot-Watt University \\
  {\tt g.h.turner@hw.ac.uk} \\\And
  Helen Hastie \\
  School of Mathematical \\ and Computer Sciences \\
  Heriot-Watt University \\
  {\tt h.hastie@hw.ac.uk} \\}

\date{}

\begin{document}
\maketitle

\begin{abstract}
Traditionally, sign language resources have been collected in controlled settings for specific tasks involving supervised sign classification or linguistic studies accompanied by specific annotation type. To date, very few who explored signing videos found online on social media platforms as well as the use of unsupervised methods applied to such resources. Due to the fact that the field is striving to achieve acceptable model performance on the data that differs from that seen during training calls for more diversity in sign language data, stepping away from the data obtained in controlled laboratory settings. Moreover, since the sign language data collection and annotation carries large overheads, it is desirable to accelerate the annotation process. Considering the aforementioned tendencies, this paper takes the side of harvesting online data in a pursuit for automatically generating and annotating sign language corpora through phoneme clustering.
\end{abstract}

\section{Introduction}
Research in sign language has focused mainly on supervised learning with the aims of classifying phonological parameters \cite{liaw2002classification,cooper2007large,buehler2009learning,buehler2010employing,cooper2012sign,7780781}, providing glosses for isolated signs \cite{1288342,ong2014sign,fagiani2015signer,devisign_attempt,mocialovtowards,8622141,tornay2019hmm}, or translation of signed utterances that consist of multiple signs to written languages \cite{neidle2012new,8614068,cihan2018neural,ko2019neural}. 

Despite the field being mainly approached with supervised methods, few attempts have been made to model sign language using unsupervised methods \cite{papapetrou2009mining,ostling2018visual} and these are mainly for data mining.

The aim of this work is to exploit sign language resources available on social media and cluster segmented phonemes without access to transcriptions during clustering. Two clustering methods are compared, one is the general DBSCAN clustering method \cite{witten2002data} and another one is iterative grouping clustering method. Experiments show that it is possible to find similar phonemes in continuous signing data using clustering approach with linguistically liable distance metric based on phonological parameters.
\\

\section{Related Work}
Deldjoo et al. \shortcite{deldjoo2016content} build a recommendation system based on the extracted visual features from the videos, such as lighting, colour, and motion. Snoek et al. \shortcite{snoek2009concept} and Hu et al. \shortcite{hu2011survey} identify colour, texture, shape, objects, and movements as features to serve as a basis for video indexing and retrieval. Furthermore, the extracted features can be grouped across the temporal dimension and used for querying similar groups during data mining. Other methods create specialised groups, such as human actions and can vary depending on the application. Unknown patterns within the groups are usually found through clustering.

Karypis et al. \shortcite{karypis2000comparison} identify bottom-up and top-down approaches to clustering. In the bottom-up approach, every element is assigned to an individual cluster and then clusters are merged in the iterative process. In the top-down approach, every element is in the same cluster and then the cluster is split into different clusters. The merging and splitting are done using a similarity metric with bottom-up approach being more common. Cluster quality can be assessed by measuring the inter-cluster entropy or relationship between the precision and recall. When no external information about the clusters is available, the inter- and intra-cluster cohesion can be used to evaluate the quality of the clusters \cite{corral2006cohesion}.

Ert{\"o}z et al. \shortcite{ertoz2003finding} describe a range of clustering methods, outlining their advantages and disadvantages and pointing out to the fact that it is difficult to design a clustering algorithm that would be able to identify non-globular clusters of different sizes in the presence of outliers without having a bias towards one or another type of clusters. Moreover, by increasing the dimensionality of the data that needs to be clustered, the similarity metric becomes more uniform, which leads to poor clustering.

Recently, Dwibedi et al. \shortcite{Dwibedi_2020_CVPR} used pre-trained ResNet \cite{he2016deep} for image recognition in an end-to-end model that generates embeddings for each frame of a video, which are then used in generating affinity matrix for the whole video. Finally, the model uses Transformer architecture \cite{vaswani2017attention} for predicting whether each frame belongs to a repetition or not. This process is more general and can be applied to any video in order to find repetitions in that video.

Despite the necessity for general purpose video clustering methods, this work extracts linguistic information from the signing videos during the pre-processing, reducing the data complexity and clustering linguistic features specific to sign languages.

\section{Data}\label{sec:data}
Four videos of different signers interpreting the song `Halo' by Beyoncé in the American Sign Language (ASL) were collected from YouTube with average time length of 4 minutes 22 seconds. The audio of the song plays in the background, the lyrics of which were used as ground truth when referring to the translation of what is being signed in the videos due to no knowledge of ASL.

\begin{figure}[H]
    \centering
    \includegraphics[width=0.25\textwidth]{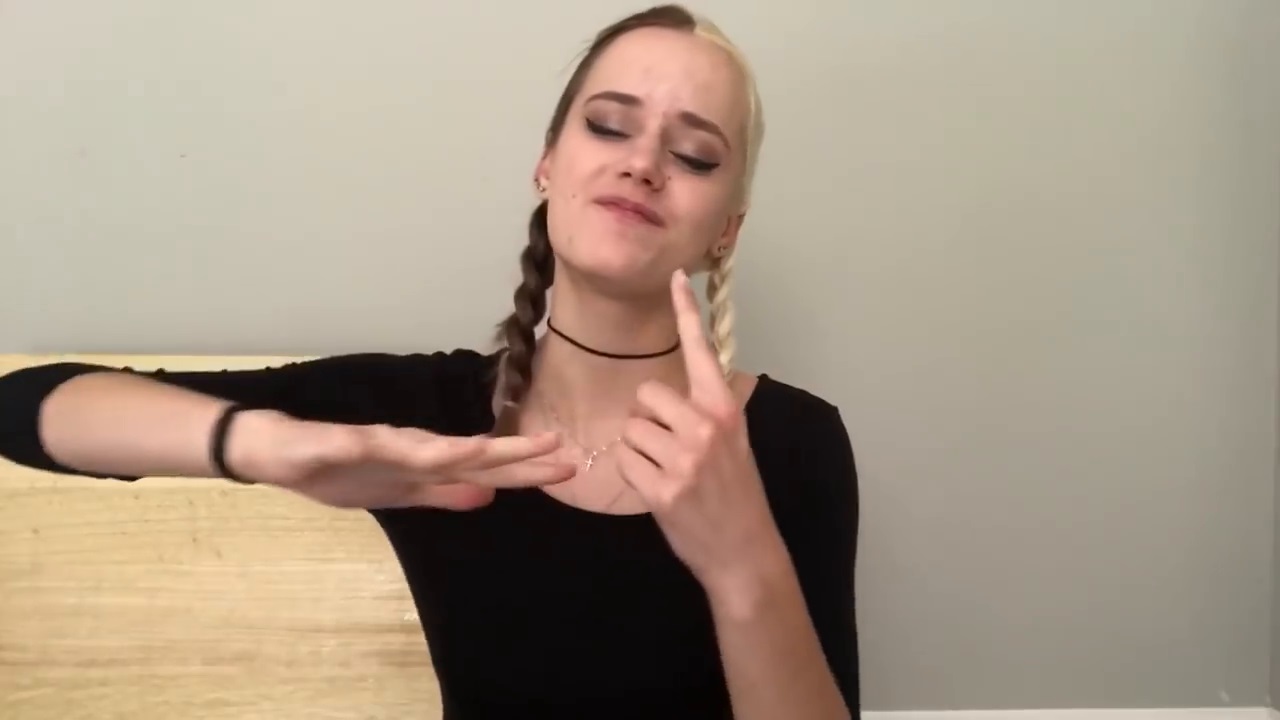}
    \includegraphics[width=0.25\textwidth]{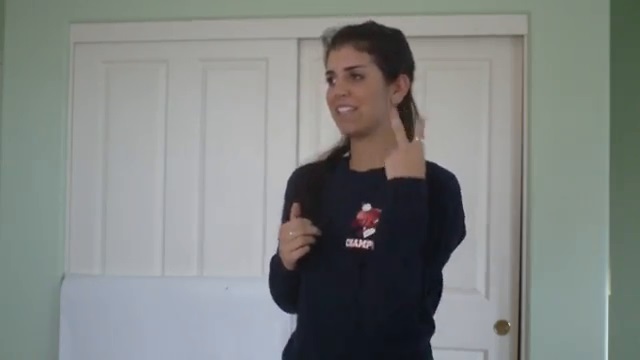}
    \\
    \includegraphics[width=0.25\textwidth]{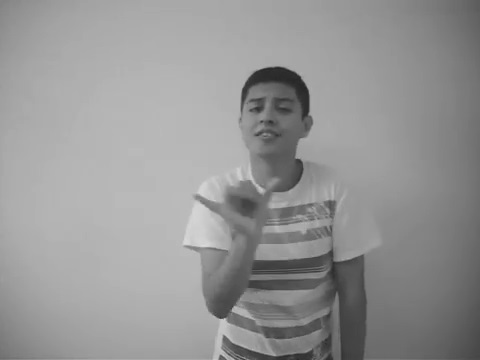}
    \includegraphics[width=0.25\textwidth]{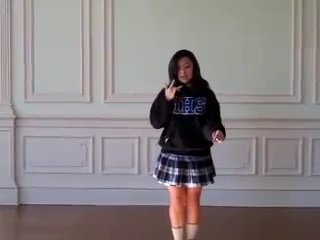}
    \caption{Screenshots from the four ASL interpreters of the song `Halo' by Beyoncé from YouTube.}
    \label{fig:data_screenshots}
\end{figure}

Figure \ref{fig:data_screenshots} shows screenshots from the four collected videos. Due to the fact that the data is collected from online resources, it is more unpredictable than the corpora as described in \cite{Schembri2013BuildingTB,troelsgaard2008electronic,crasborncorpus} that was collected in controlled environments. After a short qualitative analysis of the collected videos, the following factors were noted to vary across videos:\\
\begin{minipage}{.24\textwidth}
\vspace{1em}
\begin{itemize}
  \setlength{\itemsep}{1pt}
  \setlength{\parskip}{0pt}
  \setlength{\parsep}{0pt}
    \item Video aspect ratio
    \item Video quality
\end{itemize}
\vspace{1em}
\end{minipage}
\begin{minipage}{.24\textwidth}
\vspace{1em}
\begin{itemize}
  \setlength{\itemsep}{1pt}
  \setlength{\parskip}{0pt}
  \setlength{\parsep}{0pt}
    \item Camera proximity
    \item Background
\end{itemize}
\vspace{1em}
\end{minipage}
\begin{minipage}{.24\textwidth}
\vspace{1em}
\begin{itemize}
  \setlength{\itemsep}{1pt}
  \setlength{\parskip}{0pt}
  \setlength{\parsep}{0pt}
    \item Body orientations
    \item Dialects
\end{itemize}
\vspace{1em}
\end{minipage}
\begin{minipage}{.24\textwidth}
\vspace{1em}
\begin{itemize}
  \setlength{\itemsep}{1pt}
  \setlength{\parskip}{0pt}
  \setlength{\parsep}{0pt}
    \item Signing fluency
\end{itemize}
\vspace{1em}
\end{minipage}\\
This work takes advantage of the nature of the data since music usually has verses, which repeat themselves over a course of a song. This is useful when performing clustering since it can be assumed that every verse could belong to the same cluster.

\section{Methodology}
\begin{figure}[H]
    \centering
    \includegraphics[trim={3.5cm 22cm 3.5cm 5cm},clip,width=0.7\textwidth]{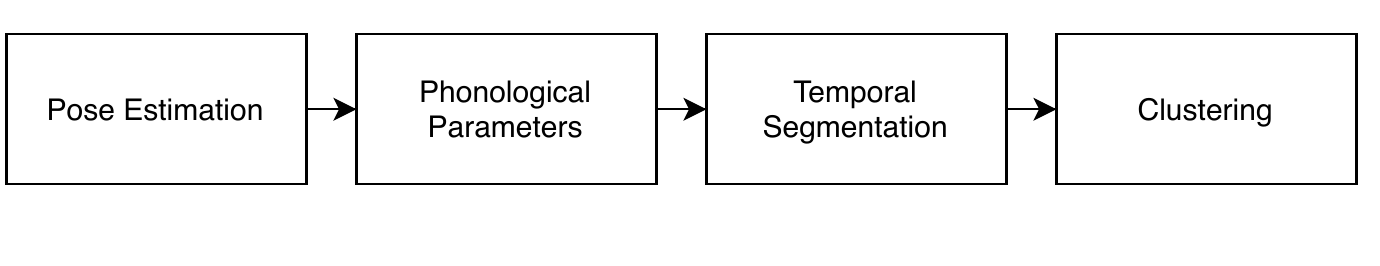}
    \caption{Methodology steps}
    \label{fig:my_label}
\end{figure}

\subsection{Pose Estimation}
We process raw signing video and estimate the pose and the hand of a signer by using the OpenPose pose estimation library \cite{wei2016cpm,simon2017hand,cao2017realtime,cao2018openpose}. The library detects 2D anatomical key-points, associated with the human body and hands in every frame. The library provides $21$ (x,y) key-points for every part of the hand and $25$ key-points for the whole body skeleton.

Due to the fact that there are variations in videos listed in Section 3, normalisation similar to Fragkiadakis et al. \shortcite{fragkiadakis-etal-2020-signing} is applied to extracted key-points to force fixed shoulder distance in all videos by resizing all \((x,y)\) positions of the inferred key-points \( (\hat{x},\hat{y})\rightarrow(x*r_x,y*r_y) \) and bring the signer to the centre of the screen by adding the difference between desired and current positions \( (x,y)\rightarrow(\hat{x}+t_x,\hat{y}+t_y) \), where \(r\) and \(t\) are the resize ratio and translation pixels. Such normalisation counteracts the fact that the signers appear at different distances away from the camera.

\subsection{Phonological Parameters}\label{sec:phonological_parameter}
Stokoe identified phonological parameters such as hand movement, hand location relative to the body, hand and extended finger orientation, and non-manual gestures such as facial expressions, all used simultaneously during signing \cite{stokoe1976dictionary}. HamNoSys \cite{hanke2004hamnosys} is one of the notation systems for sign language that has been proposed over the time to record the use of phonological parameters during signing. The notation aims to enumerate hand shape observations from all sign languages and presents combinations of palm curvature, extent of the thumb, and finger configuration to describe a handshape. As for the orientation and location phonological parameters, the orientation is split into extended finger direction and palm orientations, which can point to either of the four cardinal directions. Location, on the other hand, defines specific body parts (e.g. shoulder, head, etc.) and uses those relative body locations to identify the hand location. Movement categorises hand movement types and movement speeds. Using the key-points from the first stage, the extended finger orientation and the hand location relative to the body for both right and left hands are estimated for every frame, following the HamNoSys notation system as was reported in \cite{mocialov2020towards}. The orientation phonological parameter is divided into eight categories using trigonometry \(-\pi/2 < \arctan (q_y - p_y, q_x - p_x) < \pi/2 \) where \(q\) and \(p\) are the \((x, y)\) coordinates of the radius and middle finger metacarpal bones. Relative hand location is estimated using Euclidean distance between identified key-points and distance matrix is generated

\[
D = 
 \begin{pmatrix}
  M_r & M_l \\
  \vdots  & \vdots  \\
  N_r & N_l 
 \end{pmatrix}
\]
where \(M_{r,l} \dots N_{r,l} = \lvert q_{m \dots n}- centroid_{r,l} \rvert \) are the Euclidean distances between \( (x,y) \) body parts key-points \(q\) and \( centroid_{r,l} = (\sum_{i=1}^{N} x_{i_{r,l}} / N, \sum_{i=1}^{N} y_{i_{r,l}} / N) \) is hands centroid of all hand key-points \(N\) provided by the OpenPose library.

\subsection{Temporal Segmentation}
Using the key-points from the first stage, phonemes are segmented in a continuous signing as described in \cite{Mocialov2017TowardsCS}. The method splits continuous signing by tracking \(centroid_{r,l}\) speed \(f\) and making a split at time \(t\) when the hand movement either slows down or accelerates \(\sign(f'(t-1)) \neq \sign(f'(t))\) as it has been observed that the signing speed is different from that of the epentheses. 

\subsection{Clustering}\label{sec:clustering}
\begin{figure}[H]
\centering
    \begin{tikzpicture}
    \node[anchor=south west,inner sep=0] at (2,0) {\includegraphics[width=0.45\textwidth,trim={3cm 20cm 6cm 5cm},clip]{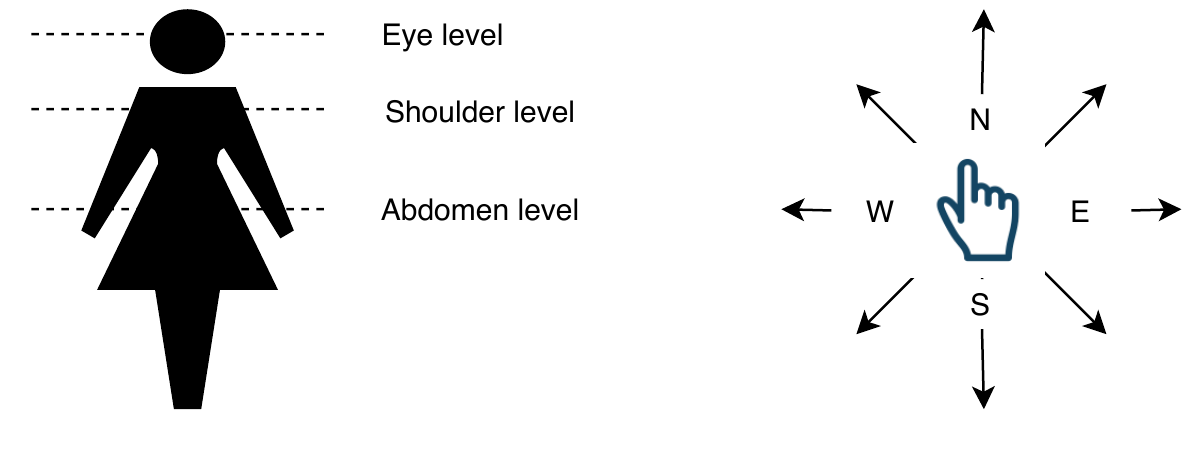}};

    \node[very thick] at (3,0) {Location Categories};
    \node[very thick] at (7,0) {Orientation Categories};
    \end{tikzpicture}
    \caption{Categories for distances for both location and orientation phonological parameters}
    \label{fig:cluster_distances}
\end{figure}

The similarity of two phonemes ($p_n \sim^{T} p_m$) is then evaluated by comparing their phonological parameters using a threshold value $T$. Both phonemes should have similar phonological parameters (\mbox{$p_n$ $\subsetneq^T$ $p_m$ $\wedge$ $p_m$ $\subsetneq^T$ $p_n$}) for the phonemes to be considered similar.

\[
\exists p_n, p_m \in P, p_n \sim^{T} p_m \iff p_n \subsetneq^T p_m \wedge p_m \subsetneq^T p_n
\]

To realise the comparison of the phonemes, the weighted Levenshtein distance \cite{rajalingam2011hierarchical} is used to measure the difference between two phonemes, which finds the minimum number of edits where an edit can be a substitution or deletion of phonological parameter and the weights determine the cost of such edits. Since the majority of clustering algorithms operate on the affinity matrices, where every item corresponds to a distance between the two items, a distance metric is needed for the orientation and location phonological parameters. The orientation phonological parameter is divided into eight equal ranges, as described in Section \ref{sec:phonological_parameter}, while the location phonological parameter is divided into three ranges (eye, shoulder, abdomen levels). Therefore, the orientation distance range is between 0 and 4 while the location distance range is between 0 and 2, as shown in Figure \ref{fig:cluster_distances}. Distances between the two items also become the weights for weighted Levenshtein distance matrix.

Since there is no way of knowing how many similar phonemes there are in a video prior to watching the video, the clustering approach selected should not assume the number of clusters. Two clustering methods will be compared. The first clustering method is an iterative grouping clustering, which iterates a list of phonemes, compares them and puts the two similar phonemes into the same cluster and repeats until all the phonemes are have been iterated. The second clustering method is the general clustering method DBSCAN. This density-based clustering method searches its neighbourhood for the members of the same cluster and can detect arbitrary-shaped cluster shapes. For the DBSCAN, the number of samples in the neighbourhood was modified \([1-5]\) for an item to be considered a core item using fixed \(\epsilon = 0.5\) since the largest distance between the two phonemes is in \([0-1]\) range. The complexity of both algorithms is $\bigO(n^2)$ \cite{schubert2017dbscan}.

\subsection{Matching Consecutive Sequences of Phonemes}
Phonemes carry little to no linguistic meaning when viewed separately as they are very short (majority of 3 frames length as shown in Figure \ref{fig:phonemes}). Therefore, the same phoneme similarity measure, as described in Section \ref{sec:clustering} is applied to a chain of consecutive phonemes in an attempt to find similar signs or even phrases in continuous signing videos. This approach works in an iterative manner. The list of phonemes is iterated and each combination of consecutive phonemes is compared to another combination of consecutive phonemes elsewhere in the video, combining consecutive phonemes if they are similar. The process is repeated until no new combinations are found. 

\section{Results}
\begin{figure}[H]
    \centering
    \includegraphics[width=0.5\textwidth]{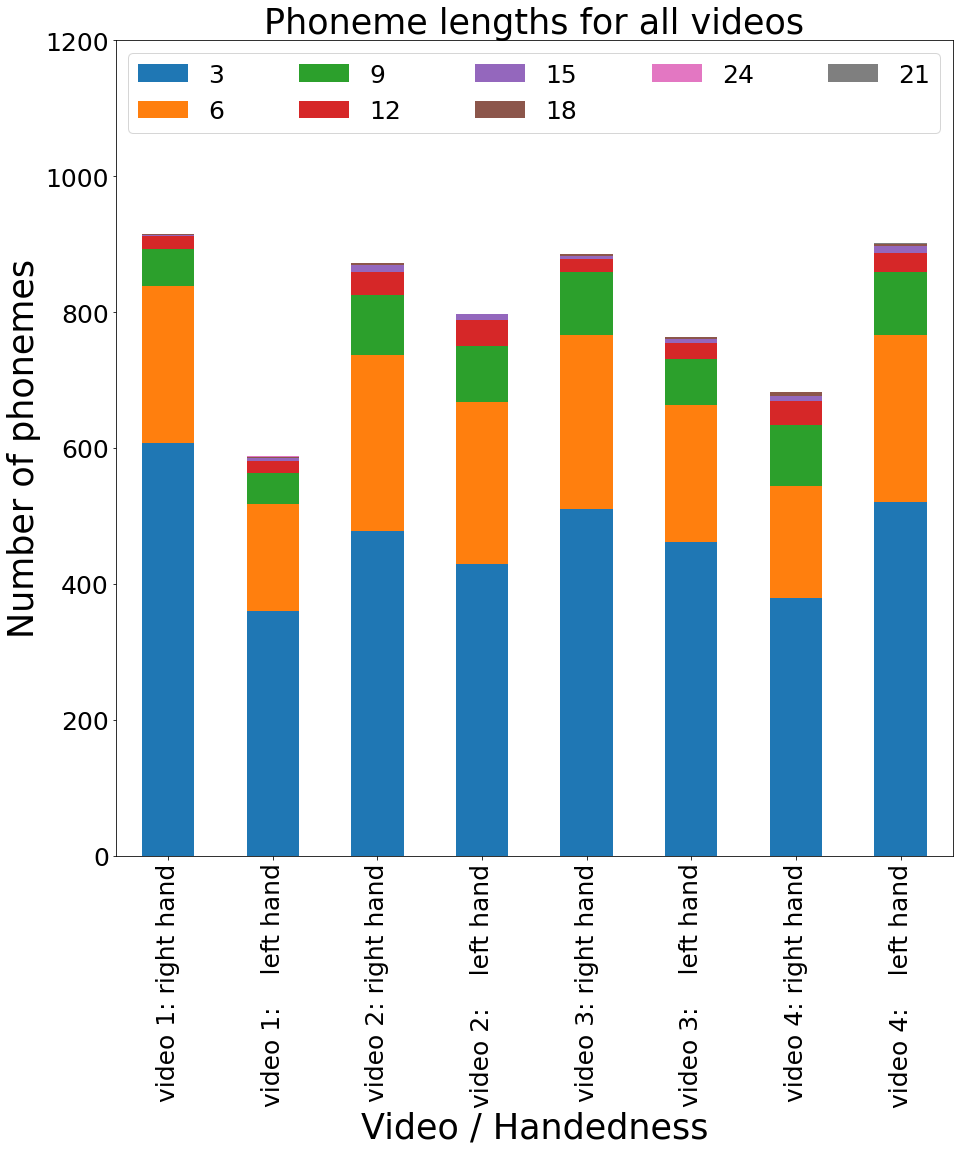}
    \caption{Total number phonemes for each video and relative phoneme sizes measured in number of frames}
    \label{fig:phonemes}
\end{figure}

Figure \ref{fig:phonemes} shows the total number of phonemes for each video, as well as, the number of phonemes of particular lengths measured in the number of frames for each video. The figure shows that using the same phoneme segmentation method on different videos produces slightly different results. For example, video 4 has fewer phonemes identified for the right hand as compared to other 3 videos. Similarly, video 1 has fewer phonemes identified for the left hand than other 3 videos. Moreover, the largest share of the overall number of phonemes belongs to the short phonemes consisting of three and six frames, while the average sign length has been reported to be of 3 seconds, or 75 frames in \cite{lichtenauer2008sign}.

\subsection{Phoneme Clustering Analysis}
This section shows how relaxing the phoneme similarity metric defined in Section \ref{sec:clustering}, affects the number of clusters and the cluster sizes with both the DBSCAN and the proposed grouping methods. 

\begin{figure}[H]
    \centering
    \includegraphics[width=0.49\textwidth]{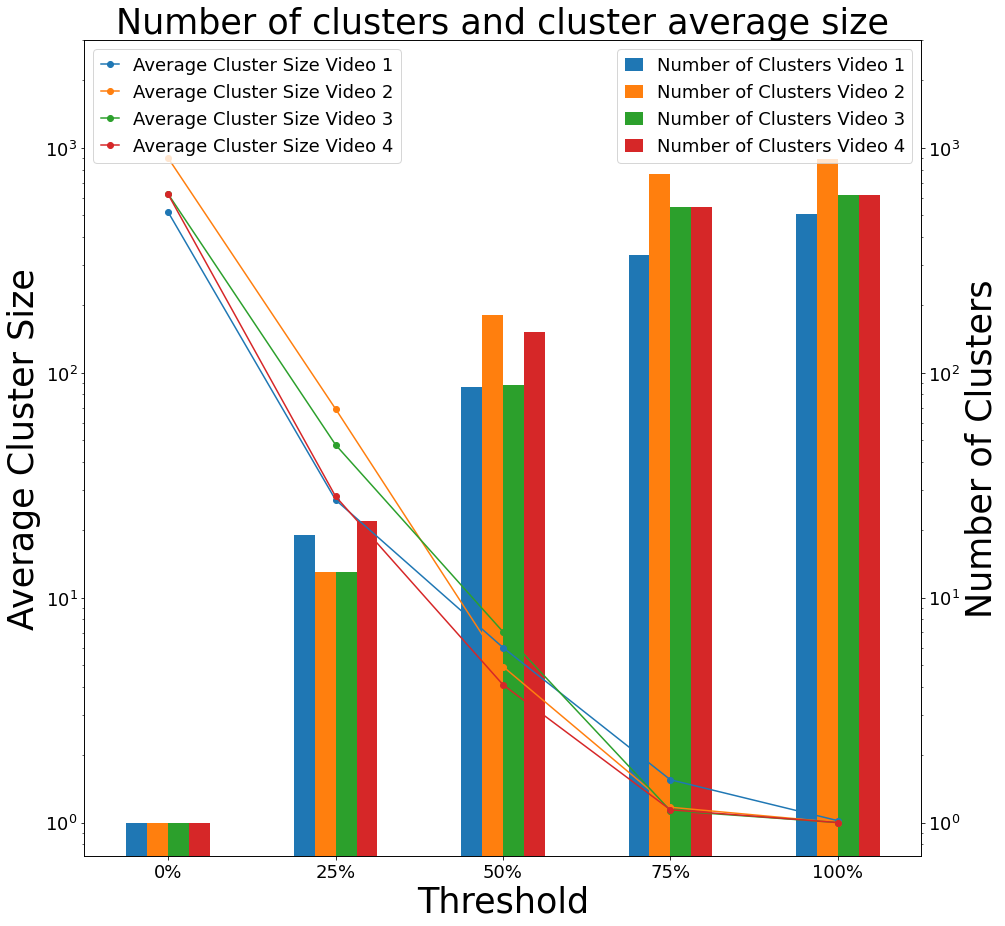}
    \includegraphics[width=0.49\textwidth]{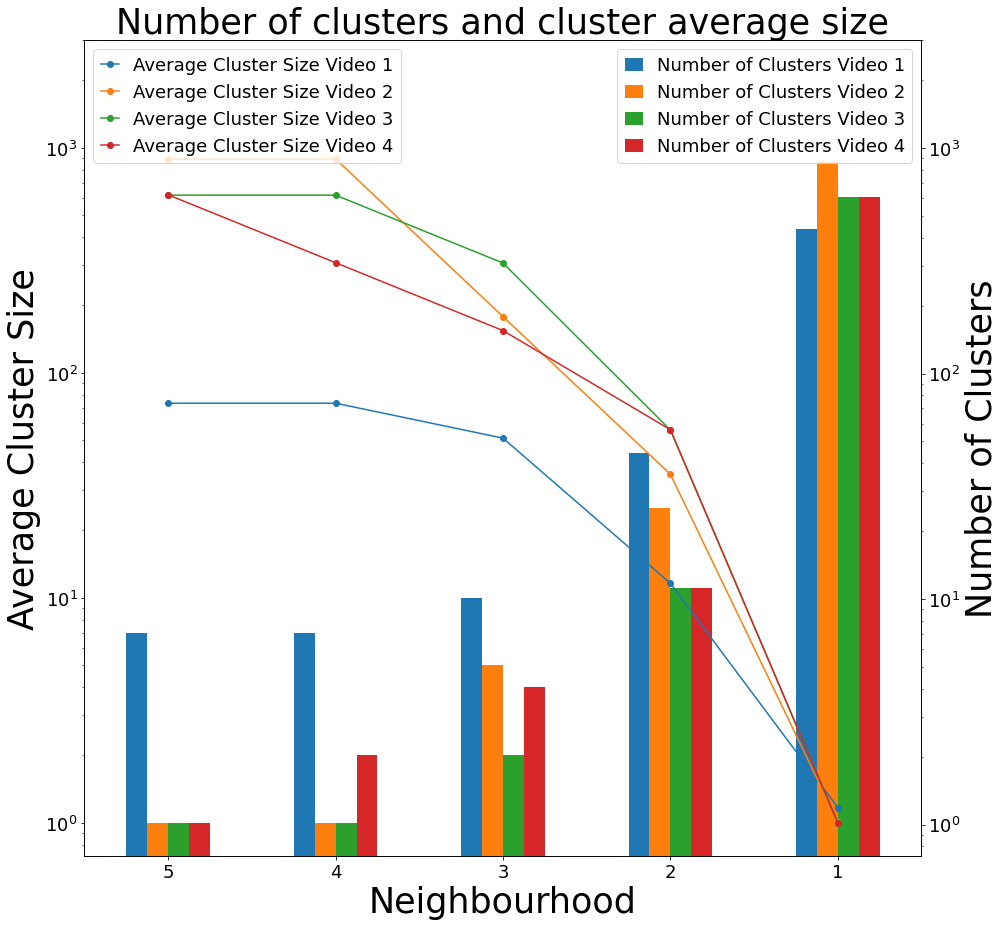}
    (a) \qquad\qquad\qquad\qquad\qquad\qquad\qquad\qquad\qquad\qquad (b)
    \caption{Number of clusters and average cluster size using grouping clustering method with different similarity thresholds (a) and using DBSCAN clustering method with different neighbourhood sizes (b).}
    \label{fig:clusters}
\end{figure}

Figure \ref{fig:clusters} (a) shows identified clusters and their average sizes with varying phoneme similarity threshold. Setting the threshold to 0\%, puts all the phonemes in one cluster, whereas setting the threshold to 100\% results in almost every phoneme being assigned to an individual cluster. This means that throughout each video, there are very few similar phonemes using proposed distance metric. This supports the fact that sign languages are fluid and the same sign can have inter-signer \cite{lucas2002location} as well as intra-signer \cite{bhuyan2006continuous} variations.

Similar behaviour is observed when clustering using the DBSCAN method, as can be seen in Figure \ref{fig:clusters} (b). As the neighbourhood size decreases, the number of clusters increase. However, it is important to notice the rate of change in the average cluster size as the similarity thresholds change (convex vs concave curve shape). Similarly, the rate of change in number of clusters is not as steep when clustered using the DBSCAN method (25\%-75\% vs 4-2 neighbours). This shows that the DBSCAN produces clusters where the number of phonemes are more uniformly distributed than with the grouping clustering method.


\subsection{Quantitative Clustering Evaluation}
Both DBSCAN and the proposed grouping clustering methods are compared using the Silhouette metric \cite{rousseeuw1987silhouettes}. The Silhouette metric has been deemed to resemble human judgement \cite{lewis2012human}. The metric is bound in range \([1,-1]\) with $1$ signifying a good clustering results and $-1$ indicating that the items in a cluster are assigned to the wrong cluster. $0$ represents many overlapping clusters in the data, which makes clustering challenging.

\begin{figure}[H]
    \centering
    \includegraphics[width=0.49\textwidth]{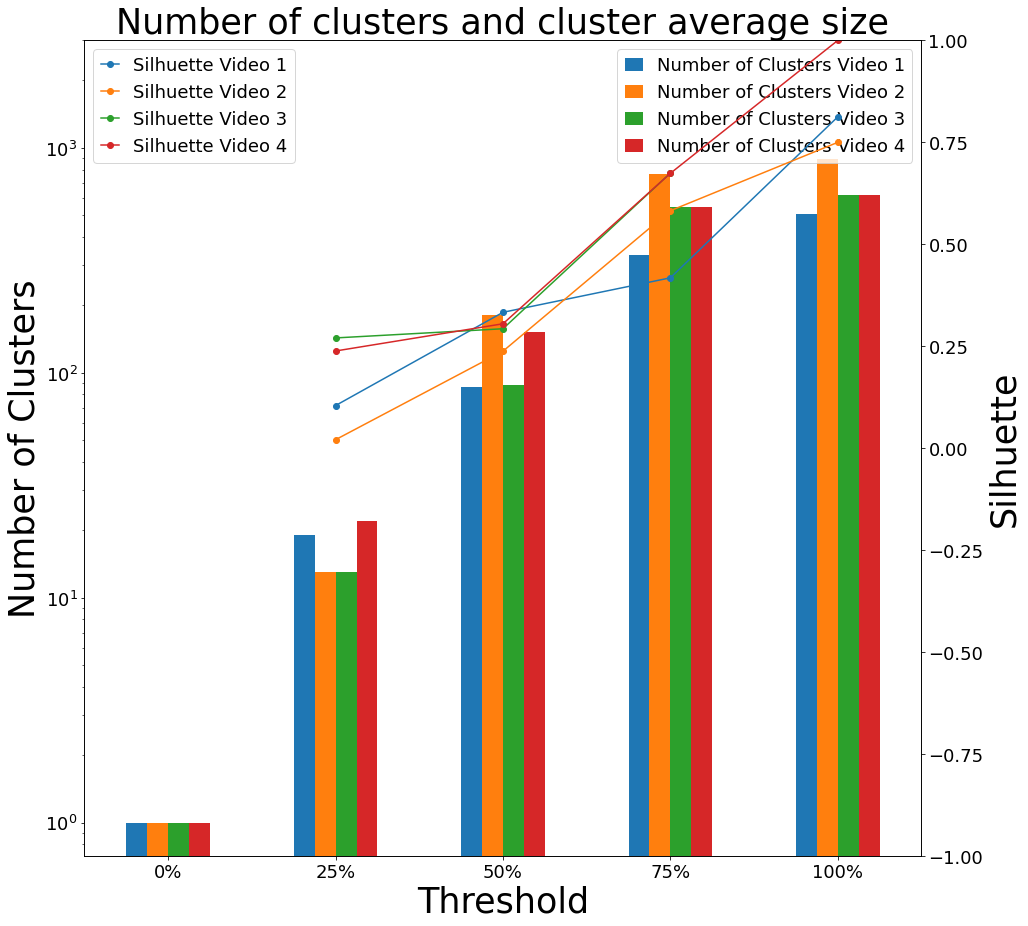}
    \includegraphics[width=0.49\textwidth]{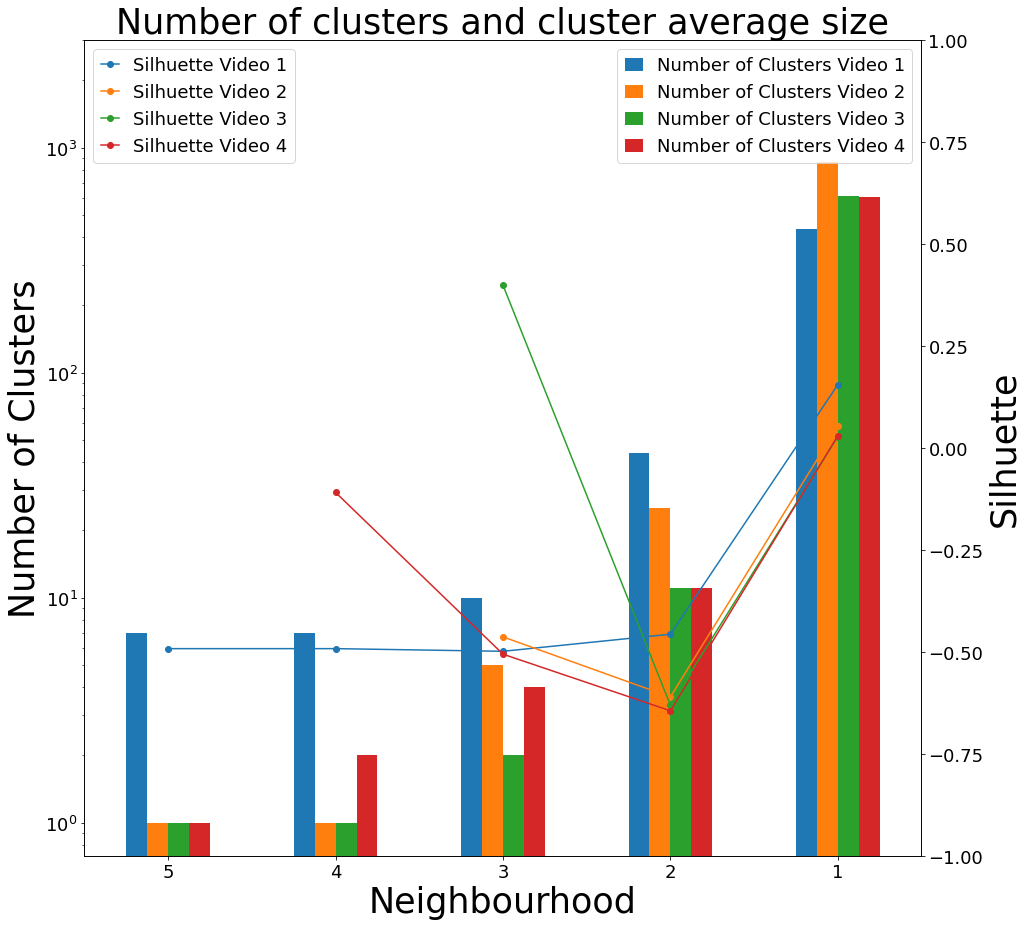}
    (a) \qquad\qquad\qquad\qquad\qquad\qquad\qquad\qquad\qquad\qquad (b)
    \caption{Silhouette clustering evaluation metric applied to grouping clustering algorithm at various similarity thresholds (a) and DBSCAN clustering algorithm with different neighbourhood sizes (b)}
    \label{fig:out_method}
\end{figure}

Figure \ref{fig:out_method} (a) shows how the Silhouette metrics changes with respect to the phoneme similarity threshold when using the proposed grouping clustering method. As the threshold is becoming more restrictive, identified clusters become clustered better at the expense of increasing number of clusters (Figure \ref{fig:clusters}). Figure \ref{fig:out_method} (b) shows DBSCAN's clustering performance with respect to the number of neighbours while having \(\epsilon=0.5\). The Silhouette coefficient is more volatile to the neighbourhood size and the number of identified clusters increases as the neighbourhood becomes smaller.

What is worth noting is that for every video, the proposed grouping clustering method has relatively similar Silhouette value while DBSCAN at \mbox{$neighbourhood=2$} and \mbox{$neighbourhood=3$} has very different Silhouette values for each video, which makes it unpredictable to use in the general case. On the other hand, having more uniformly distributed cluster sizes at different thresholds and relatively good Silhouette metric makes it an attractive clustering option for sign language data.

\subsection{Qualitative Clustering Evaluation}
This section shows example phonemes from different clusters when clustered using the grouping method with the similarity threshold set to 50\%. 

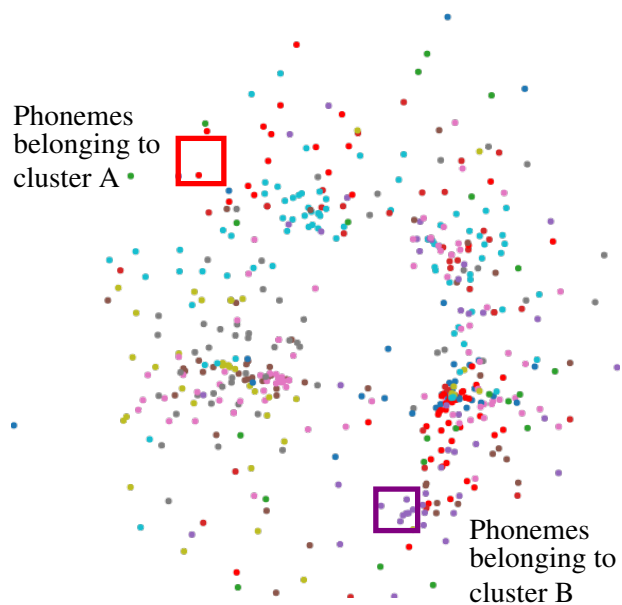
\begin{figure}[H]
\centering
    \begin{tikzpicture}
    \node[anchor=south west,inner sep=0] at (2,0) {\includegraphics[width=0.5\textwidth,trim={4cm 4cm 4cm 4cm},clip]{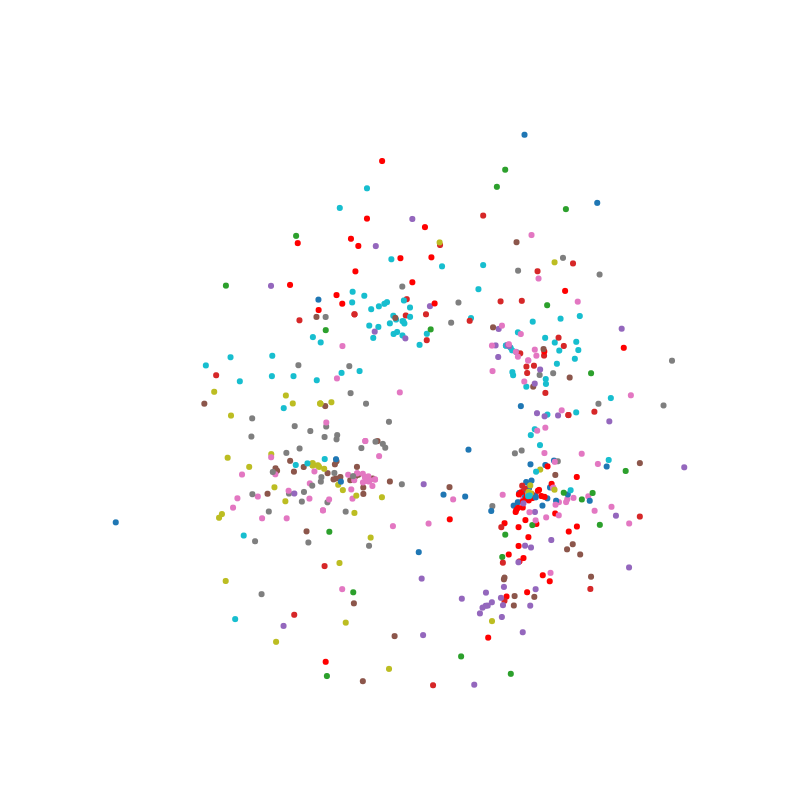}};
    \draw[draw=red,line width=0.7mm] (4.2,5.5) rectangle ++(0.6,0.6);
    \node at (3,6) {\shortstack[l]{Phonemes\\belonging to\\cluster A}};
    \draw[draw=violet,line width=0.7mm] (6.8,0.9) rectangle ++(0.55,0.55);
    \node at (9,0.5) {\shortstack[l]{Phonemes\\belonging to\\cluster B}};
    \end{tikzpicture}
    \caption{Single video phonemes clustered using iterative grouping method with 50\% threshold where each cluster has a distinct colour}
    \label{fig:all_clusters}
\end{figure}

Figure \ref{fig:all_clusters} shows visualisation of all the clusters for a single video when plotting two principal components on (x,y) axes. Clusters are identified using the iterative grouping method with 50\% threshold. Figures below show two examples for each region of interest (red and violet).

\begin{figure}[H]
    \centering
    \shortstack[c]{\vspace{0.7cm}(a)} \includegraphics[width=0.13\textwidth,trim={5cm 1cm 5cm 1cm},clip]{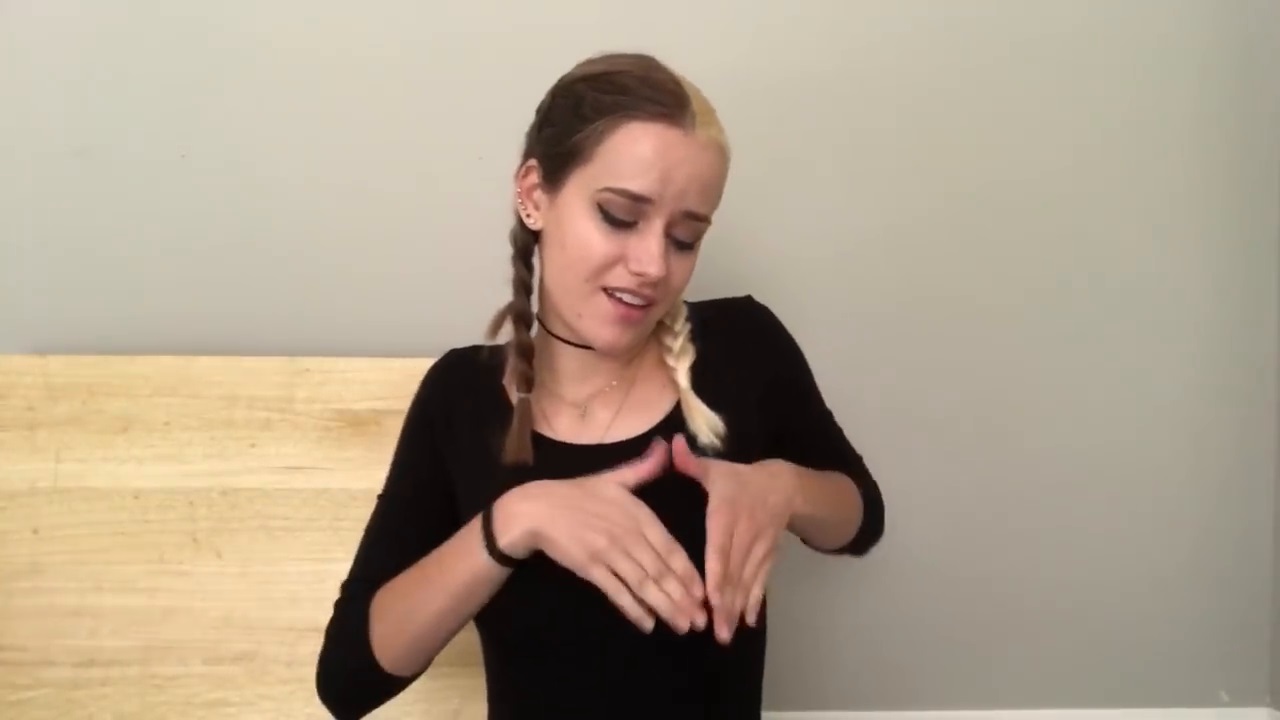}\hspace{-1em}
    \includegraphics[width=0.13\textwidth,trim={5cm 1cm 5cm 1cm},clip]{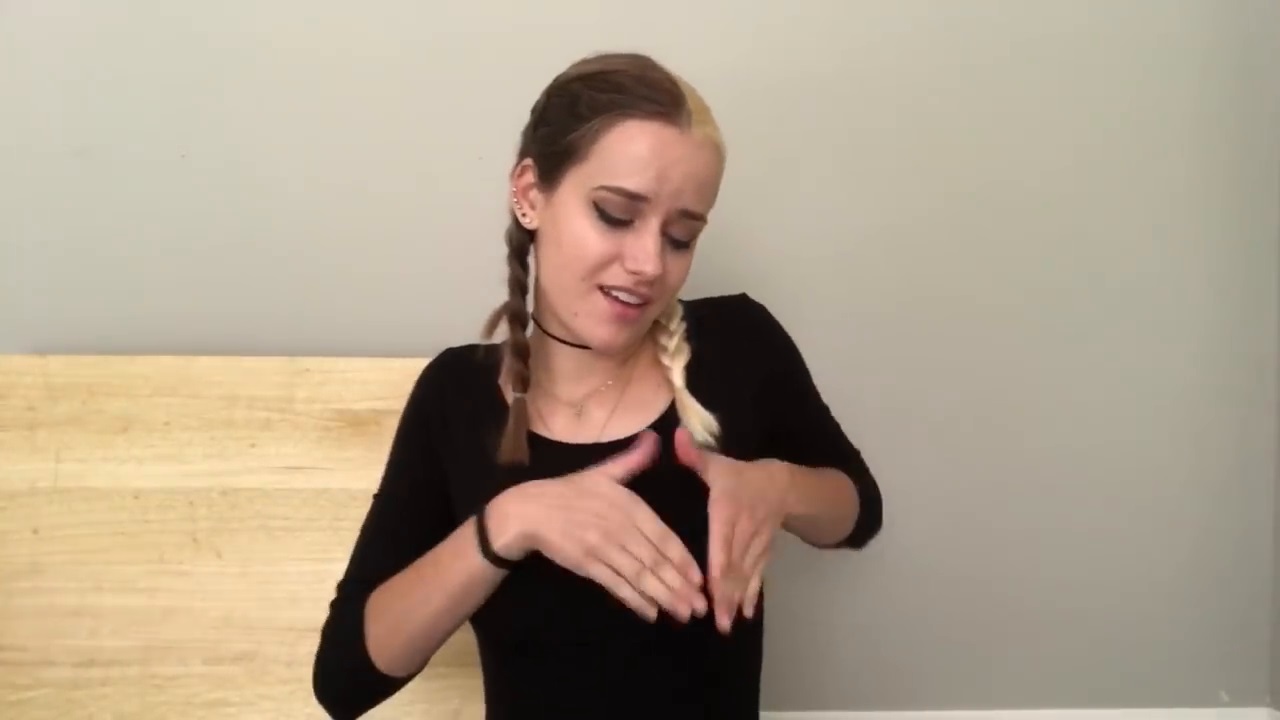}\hspace{-1em}
    \includegraphics[width=0.13\textwidth,trim={5cm 1cm 5cm 1cm},clip]{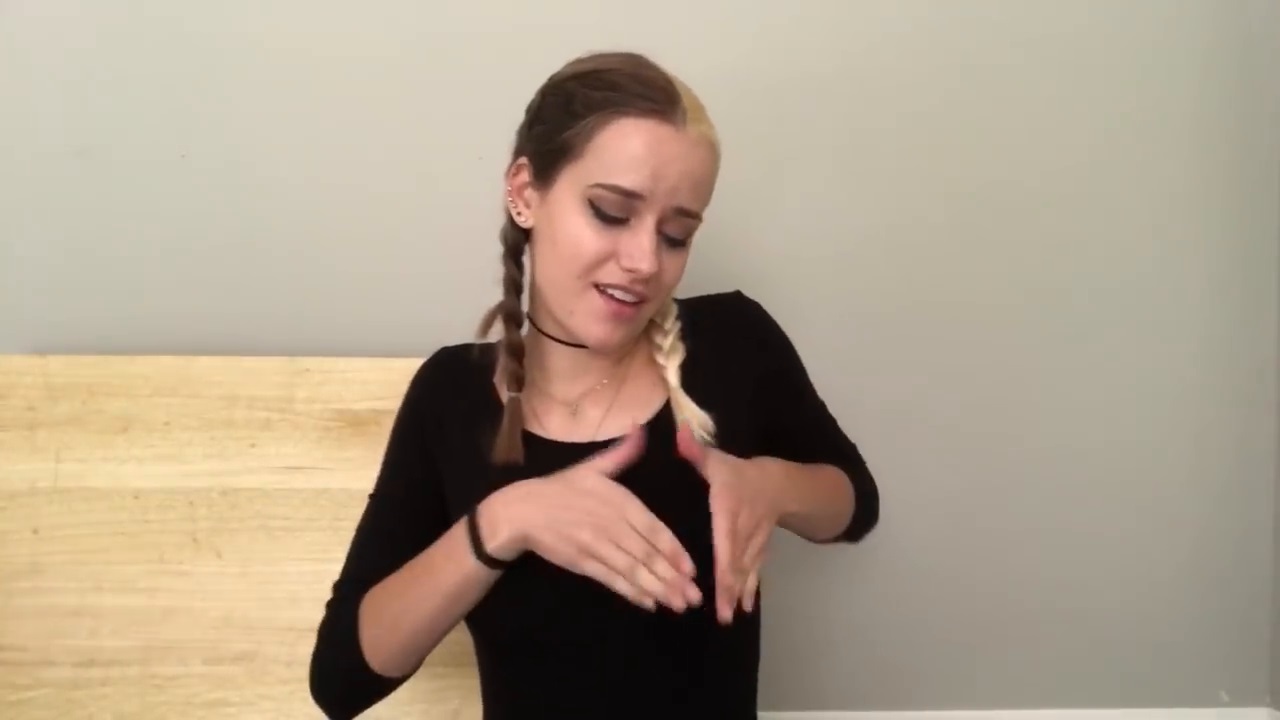}\hspace{-1em}
    \includegraphics[width=0.13\textwidth,trim={5cm 1cm 5cm 1cm},clip]{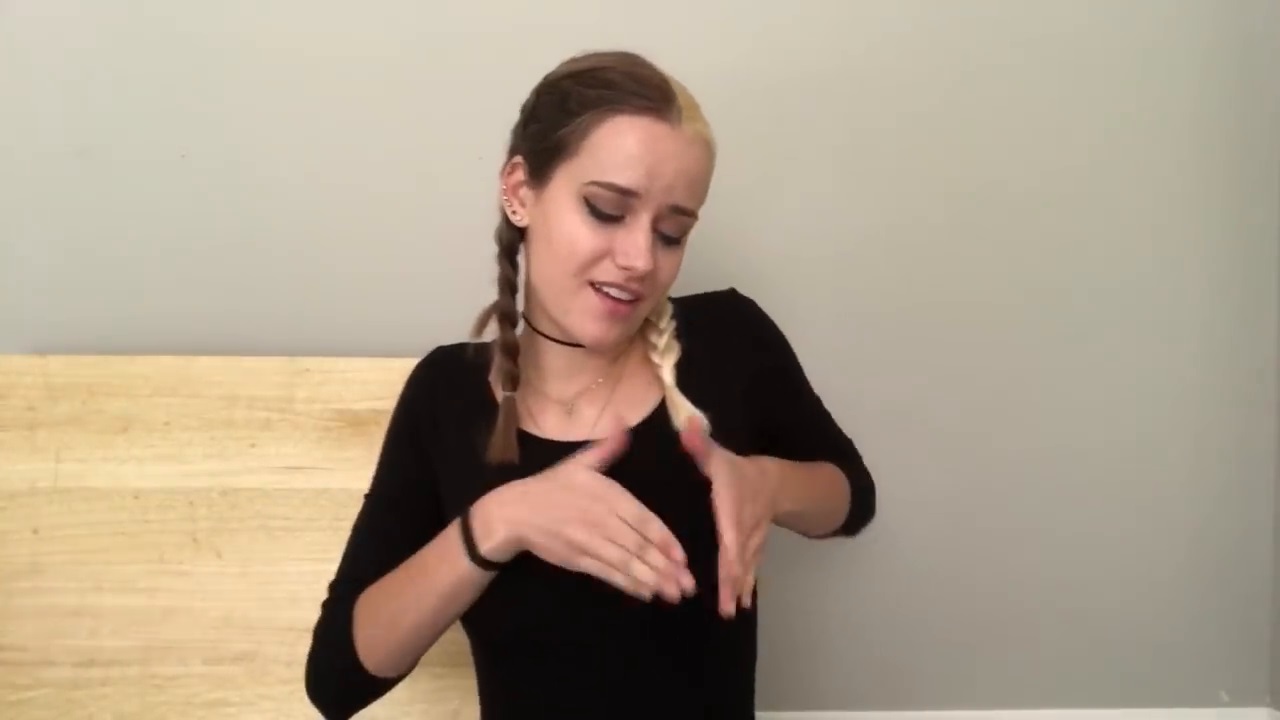}\hspace{-1em}
    \includegraphics[width=0.13\textwidth,trim={5cm 1cm 5cm 1cm},clip]{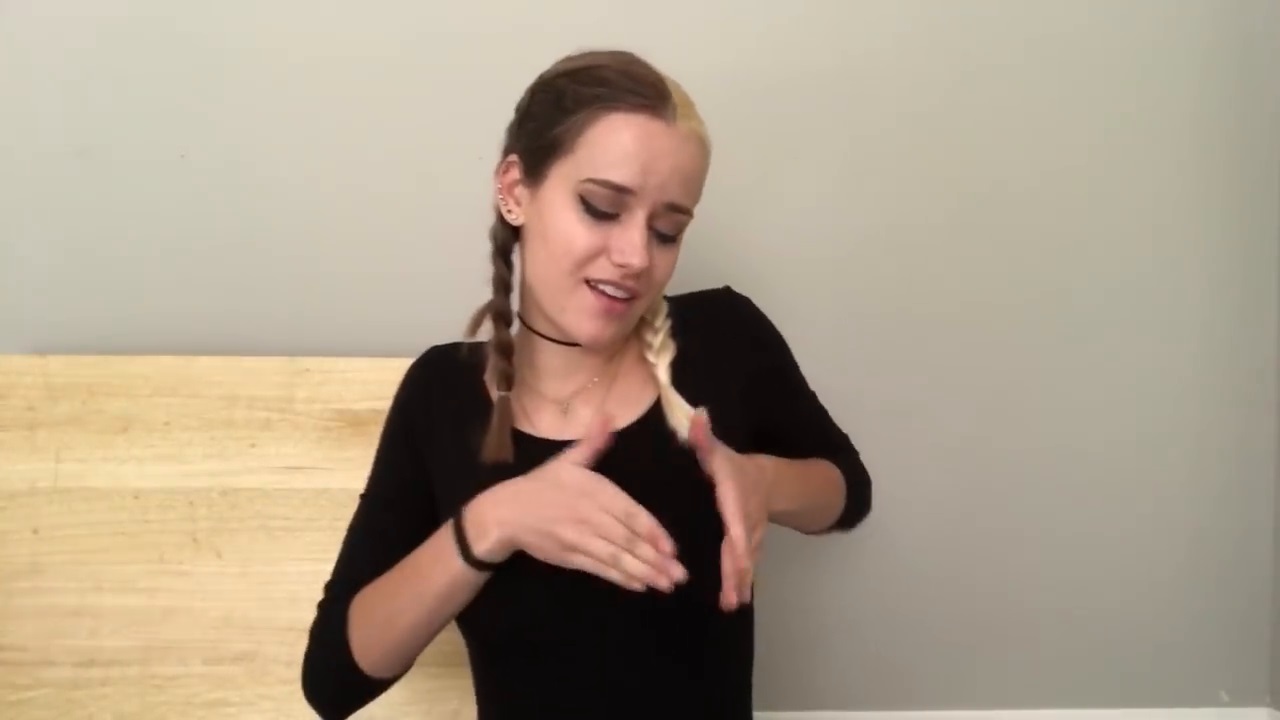}\hspace{-1em}
    \includegraphics[width=0.13\textwidth,trim={5cm 1cm 5cm 1cm},clip]{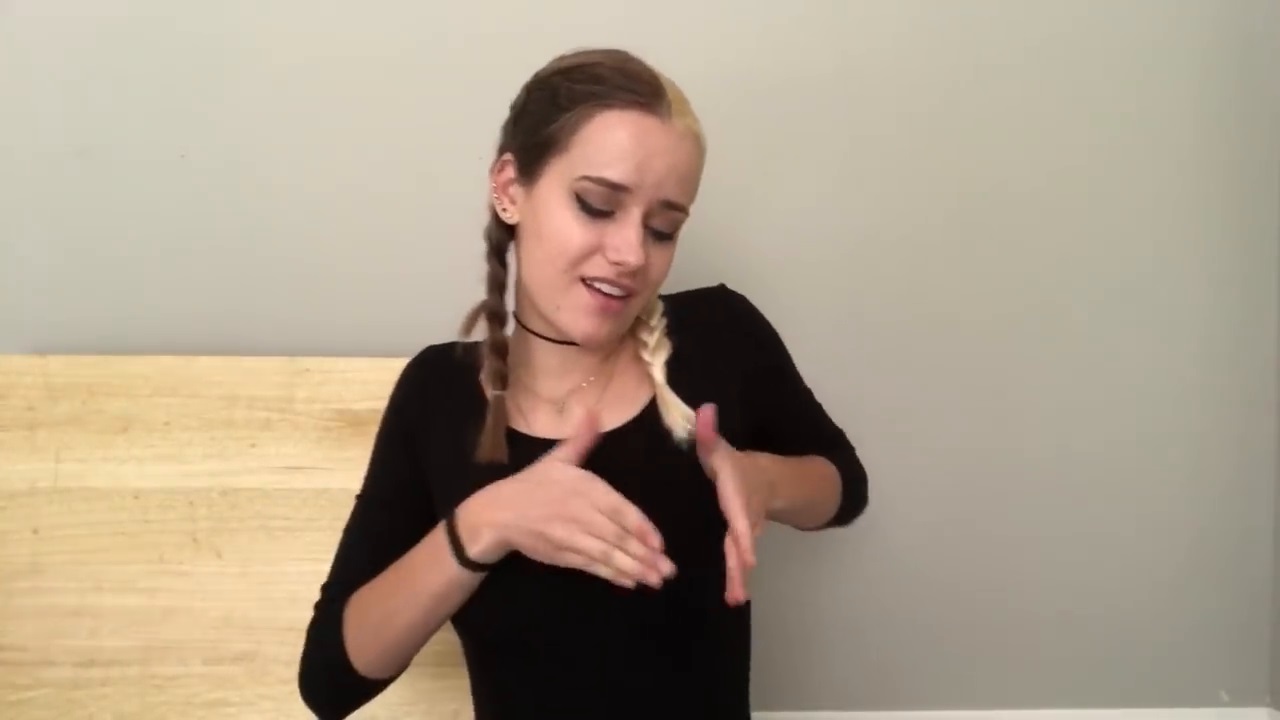}\hspace{-1em}
    \includegraphics[width=0.13\textwidth,trim={5cm 1cm 5cm 1cm},clip]{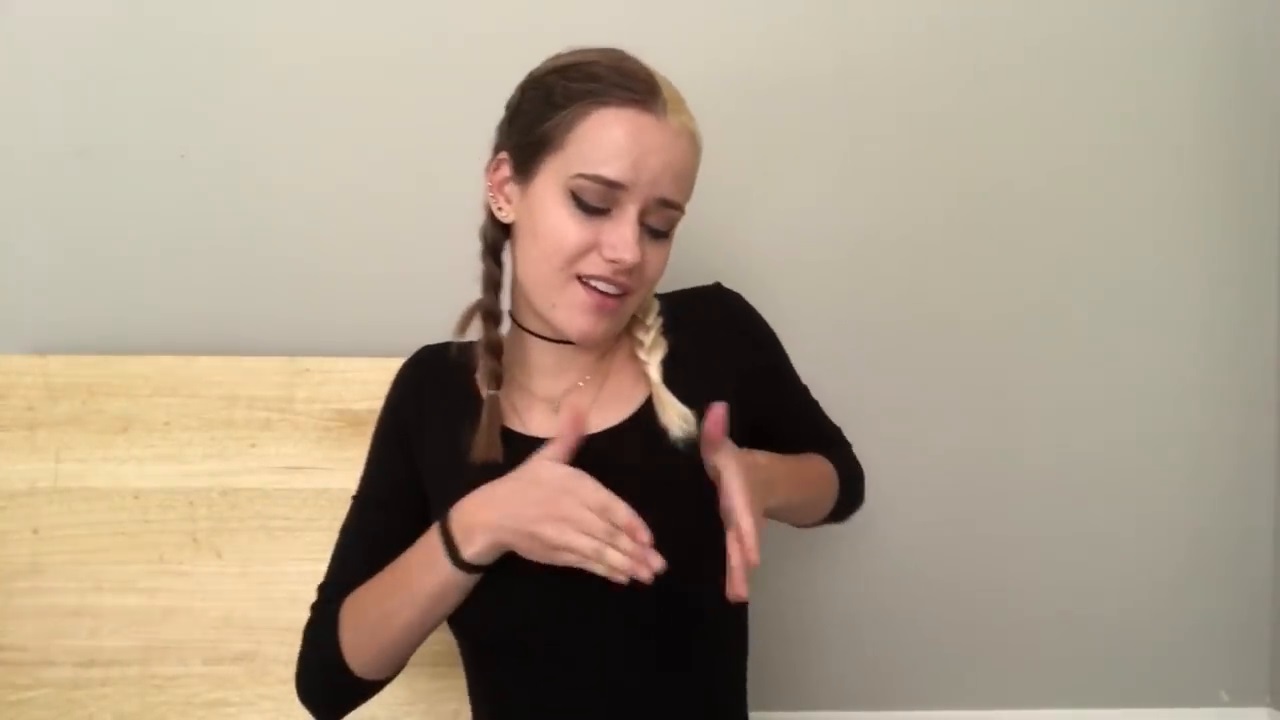}
    
    \shortstack[c]{\vspace{0.7cm}(b)} \includegraphics[width=0.13\textwidth,trim={5cm 1cm 5cm 1cm},clip]{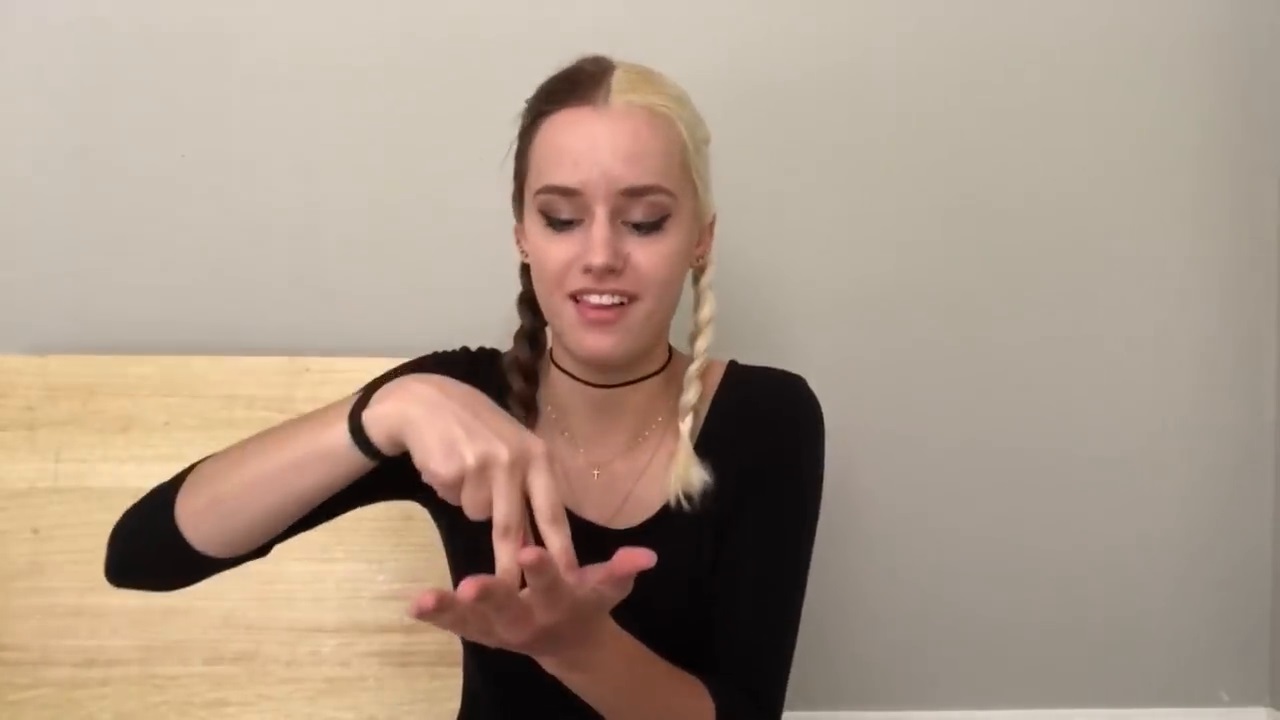}\hspace{-1em}
    \includegraphics[width=0.13\textwidth,trim={5cm 1cm 5cm 1cm},clip]{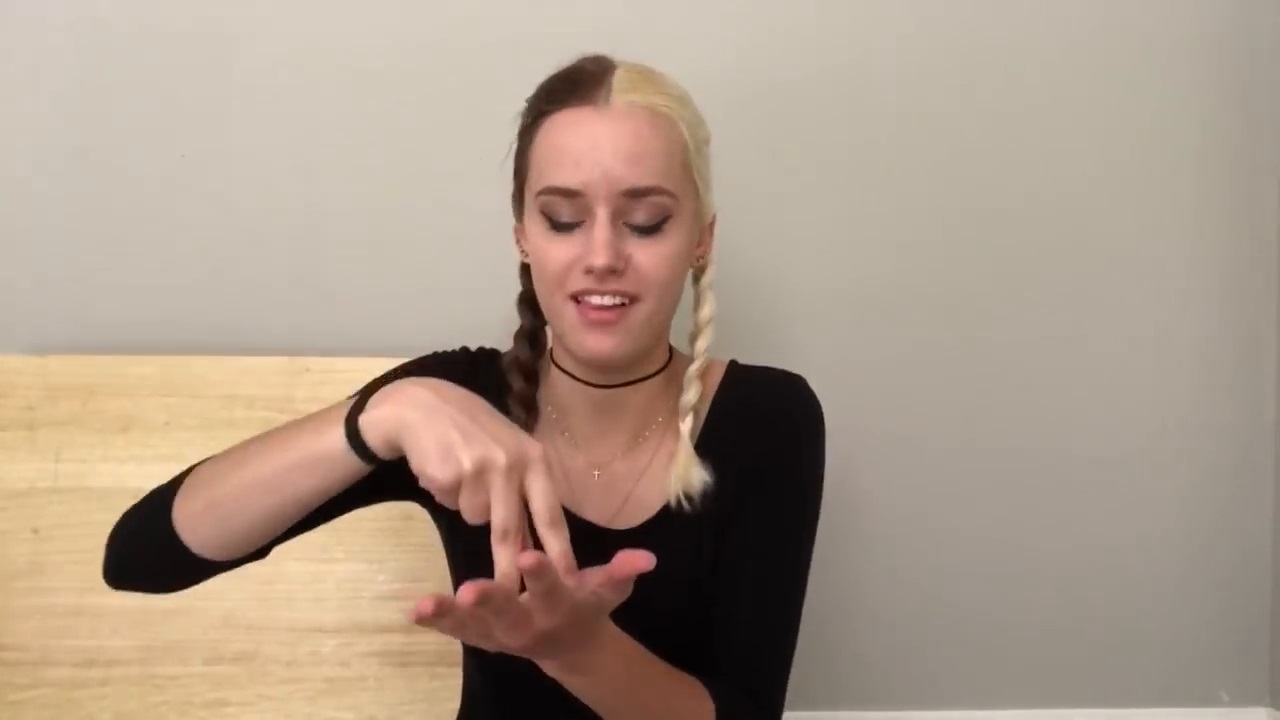}\hspace{-1em}
    \includegraphics[width=0.13\textwidth,trim={5cm 1cm 5cm 1cm},clip]{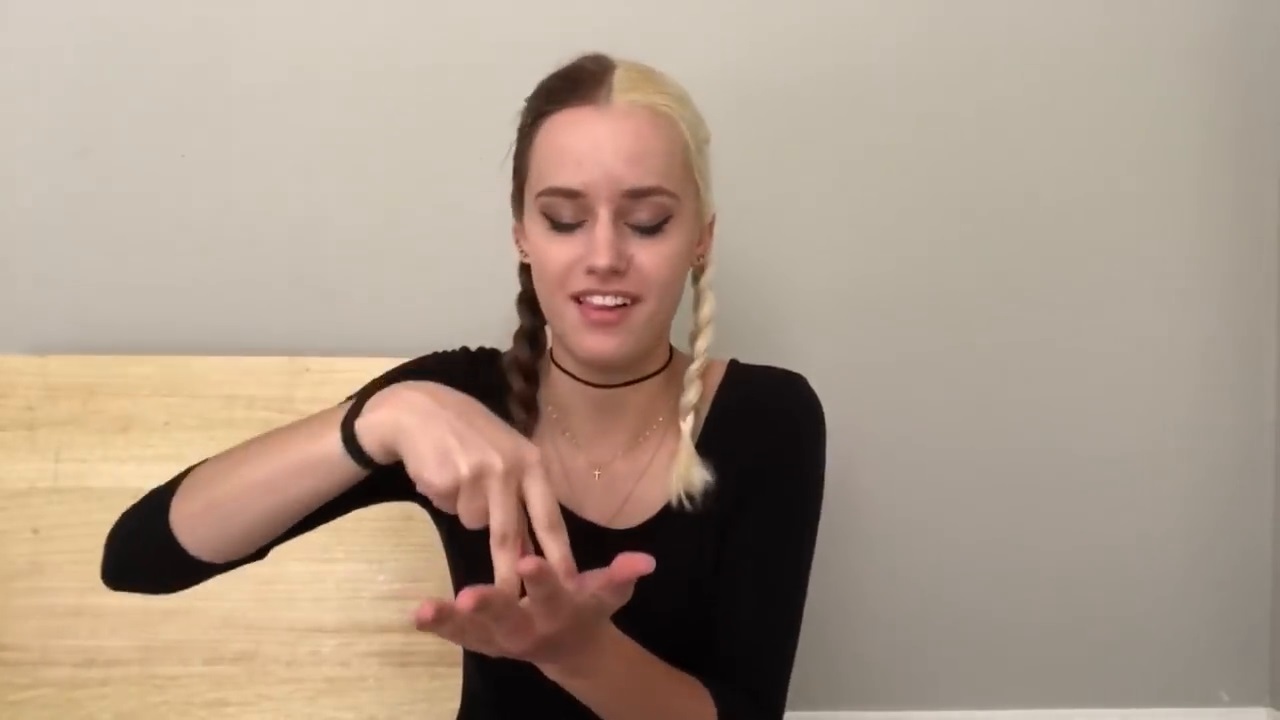}\hspace{-1em}
    \includegraphics[width=0.13\textwidth,trim={5cm 1cm 5cm 1cm},clip]{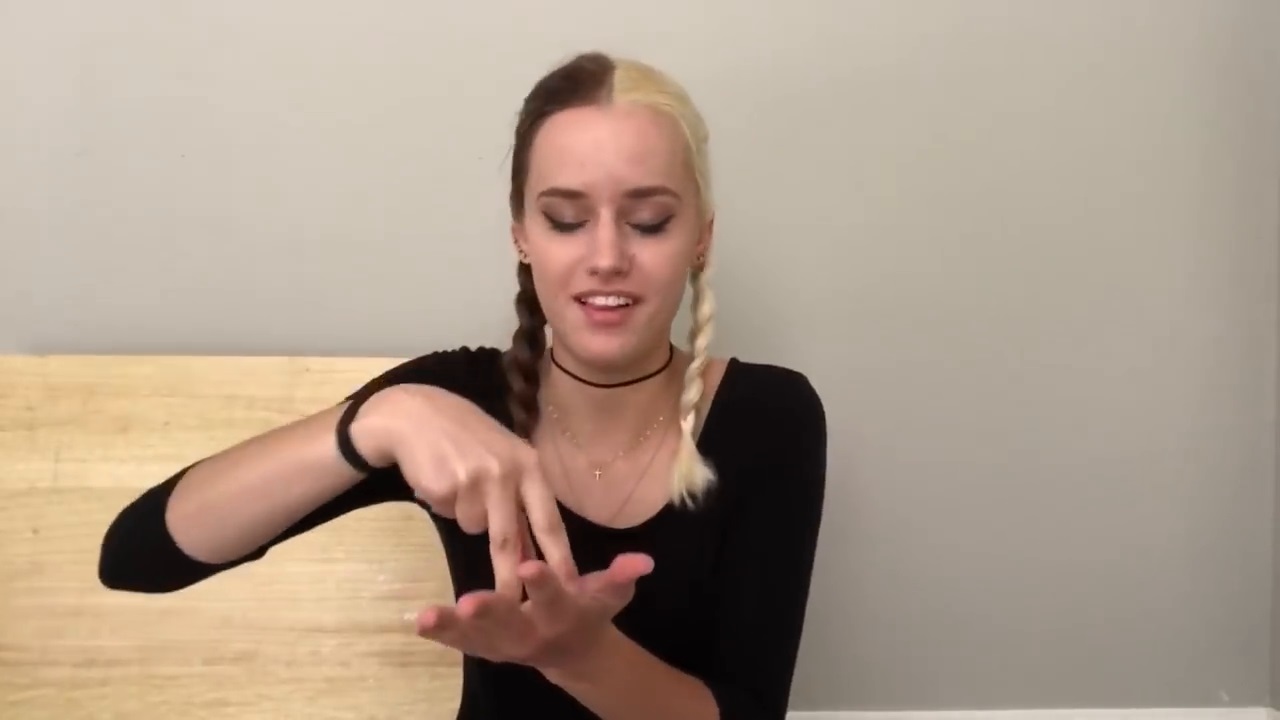}\hspace{-1em}
    \includegraphics[width=0.13\textwidth,trim={5cm 1cm 5cm 1cm},clip]{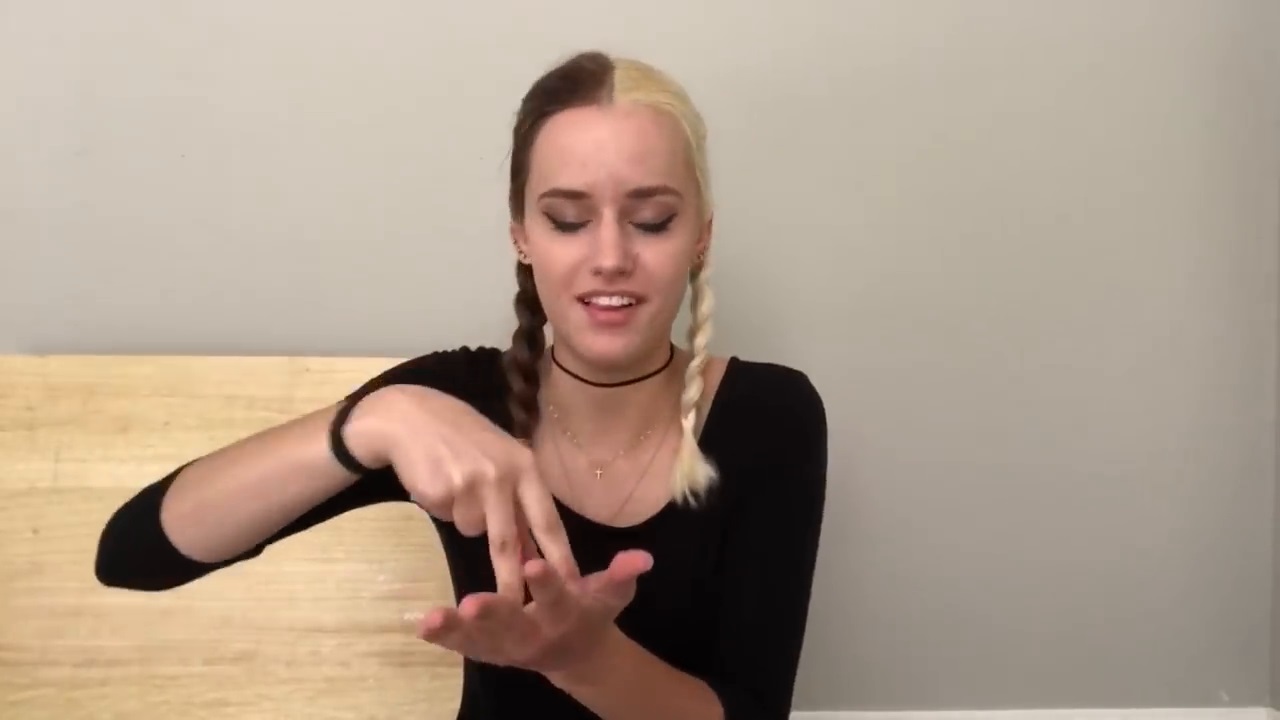}\hspace{-1em}
    \includegraphics[width=0.13\textwidth,trim={5cm 1cm 5cm 1cm},clip]{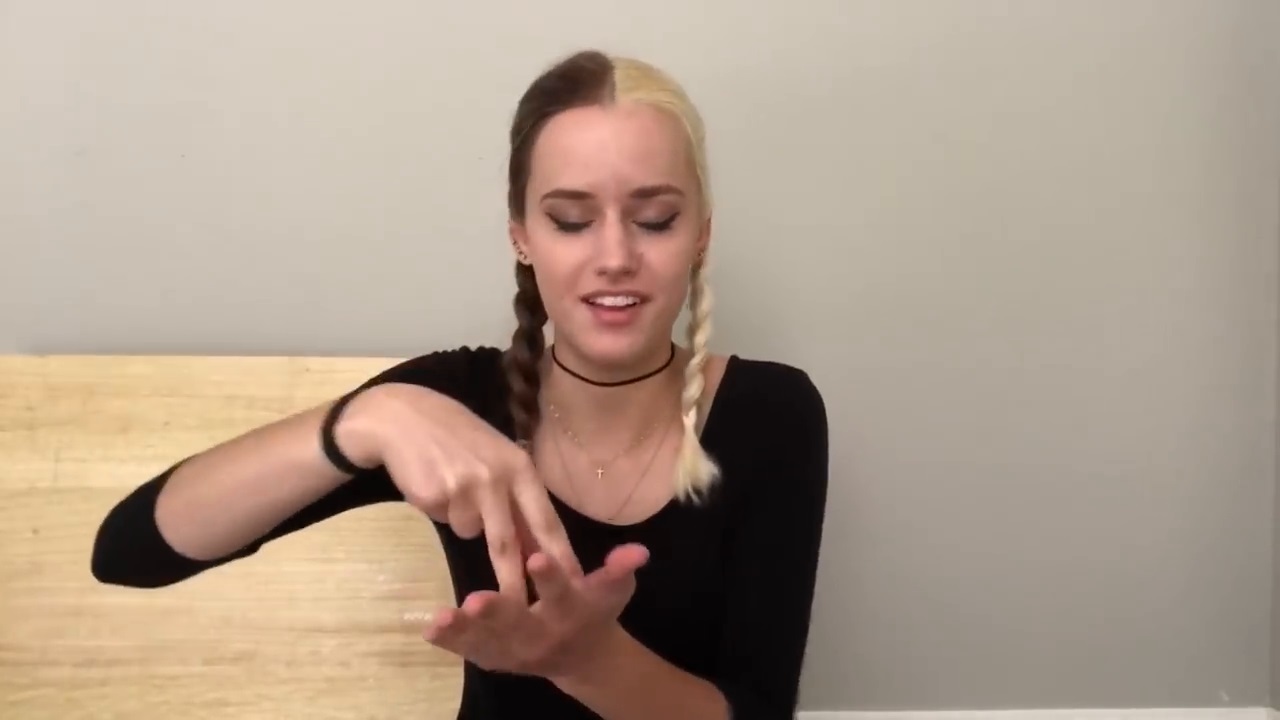}\hspace{-1em}
    \includegraphics[width=0.13\textwidth,trim={5cm 1cm 5cm 1cm},clip]{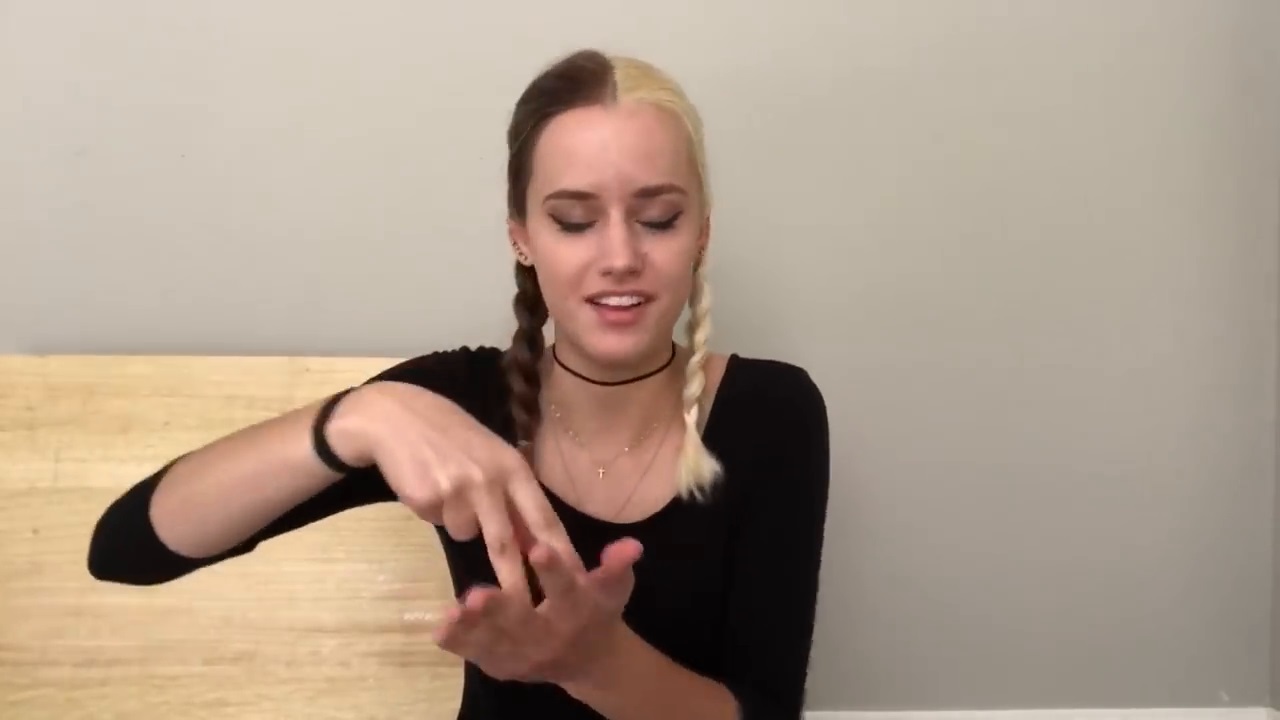}
    \caption{Two similar phonemes (Frames 1196-1202 for the phoneme (a) \& 3556-3562 for the phoneme (b)) from cluster A (red). Phoneme (a) being a part of the `I found a way to let you in' lyrics and phoneme (b) being a part of the `to pull me back to the ground again' lyrics.}
    \label{fig:visual1}
\end{figure}

\begin{figure}[H]
    \centering
    \shortstack[c]{\vspace{0.7cm}(a)} \includegraphics[width=0.13\textwidth,trim={5cm 1cm 5cm 1cm},clip]{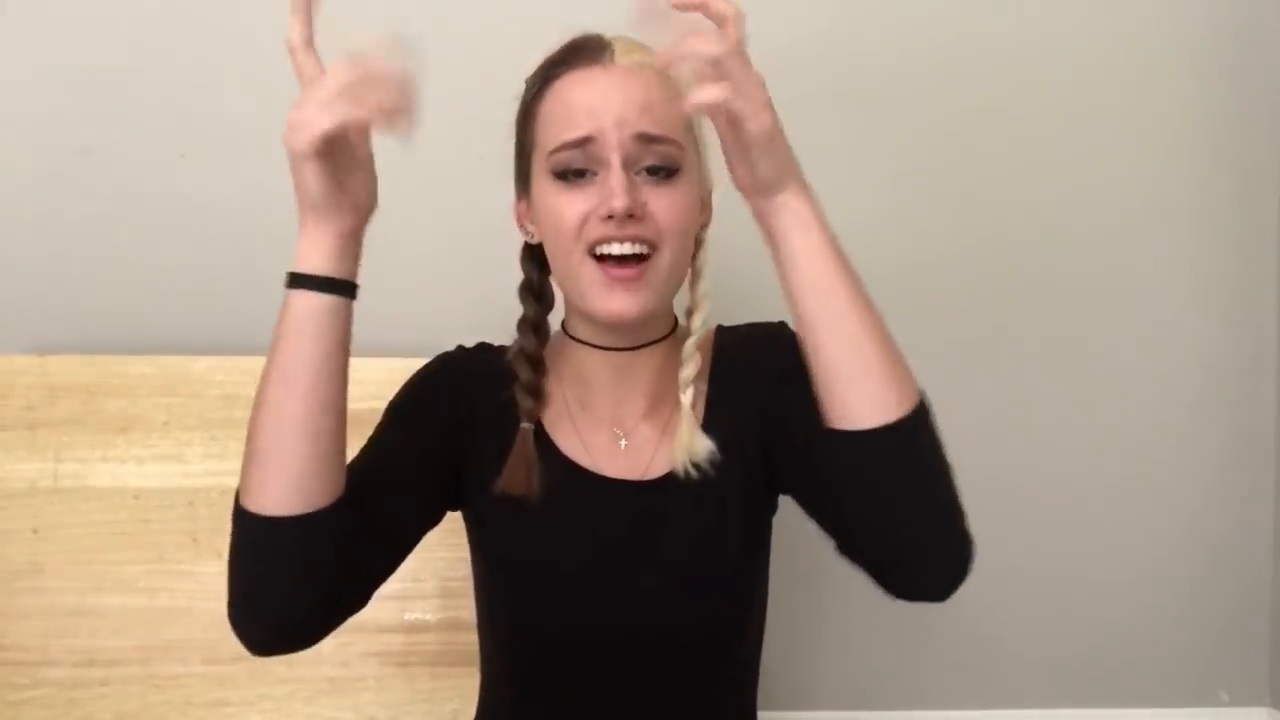}\hspace{-1em}
    \includegraphics[width=0.13\textwidth,trim={5cm 1cm 5cm 1cm},clip]{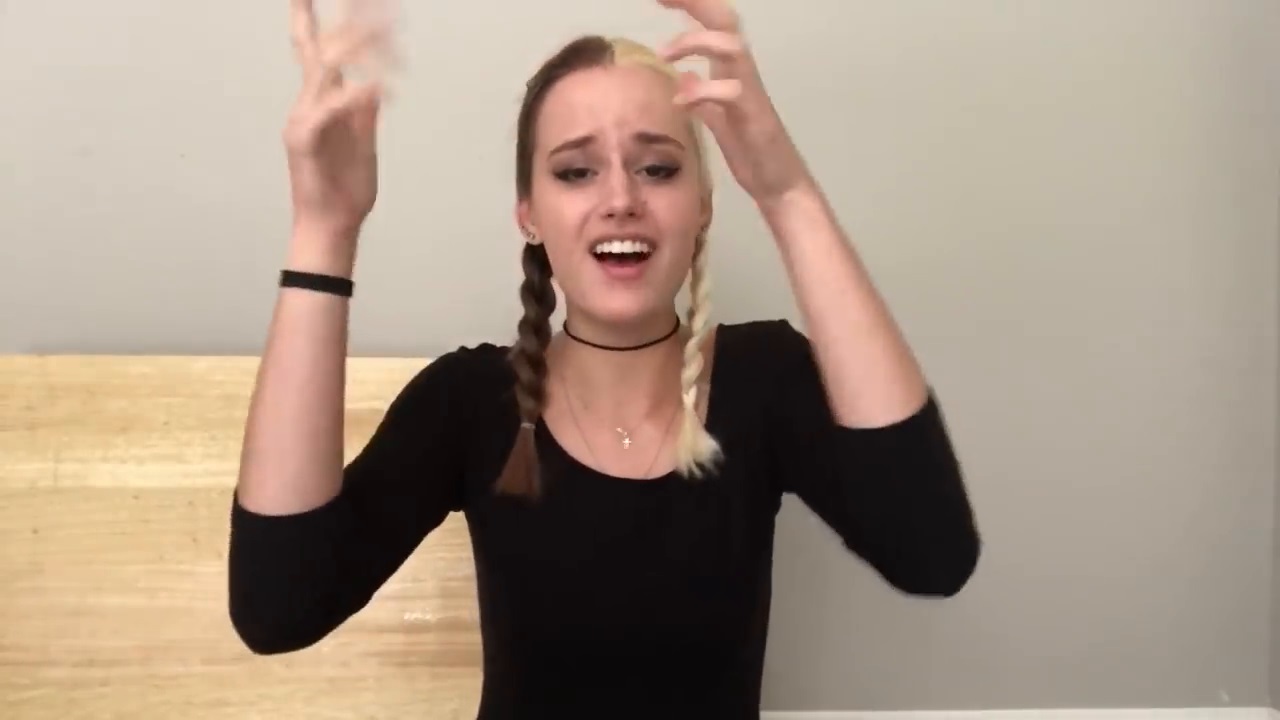}\hspace{-1em}
    \includegraphics[width=0.13\textwidth,trim={5cm 1cm 5cm 1cm},clip]{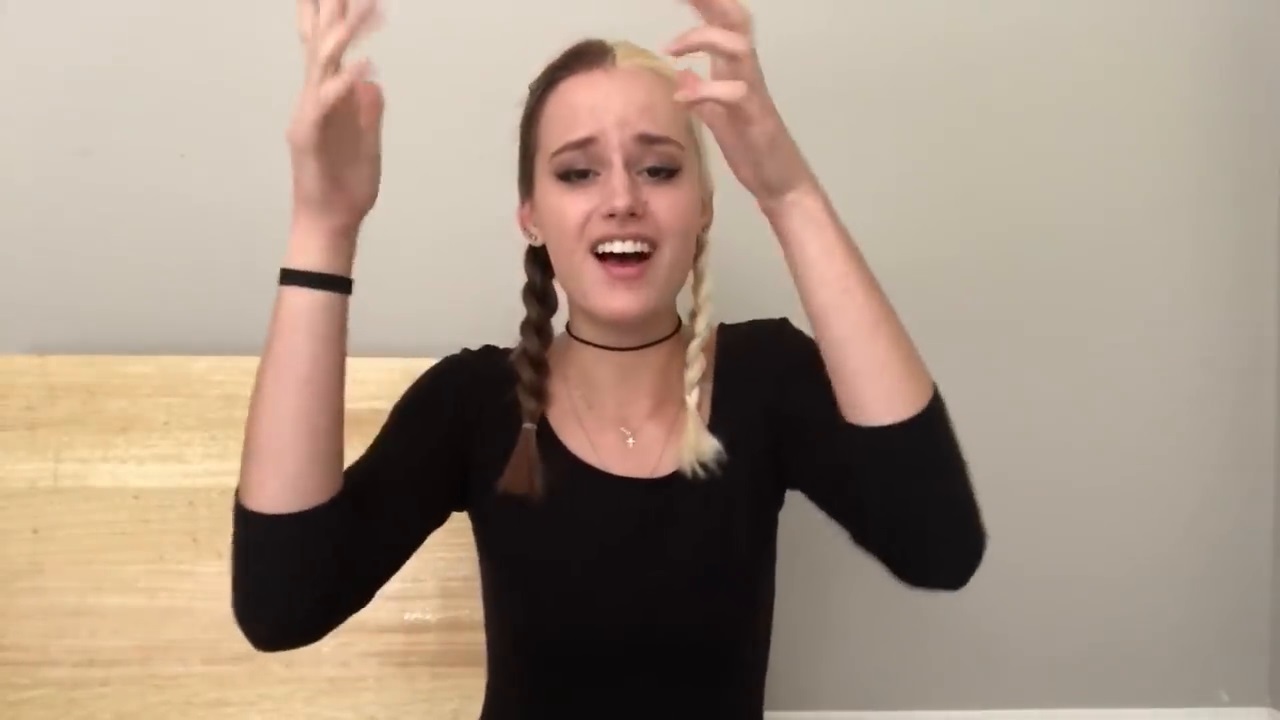}\hspace{-1em}
    \includegraphics[width=0.13\textwidth,trim={5cm 1cm 5cm 1cm},clip]{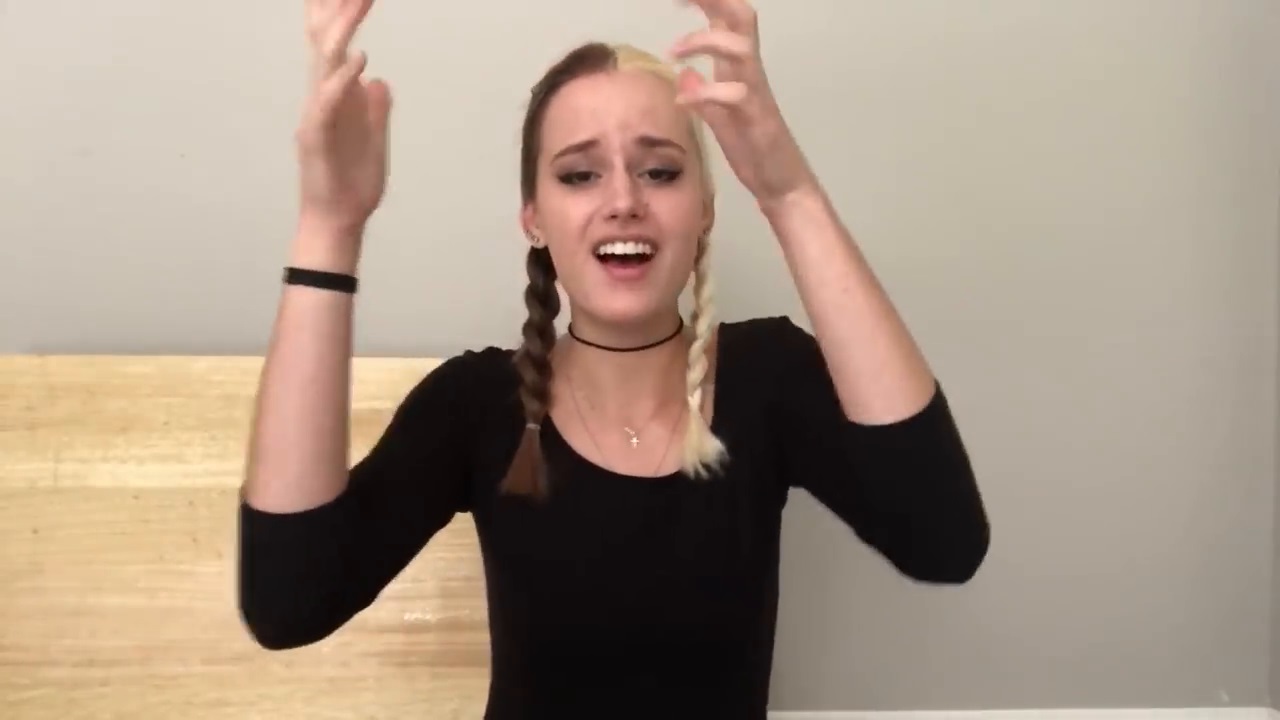}\hspace{-1em}
    \includegraphics[width=0.13\textwidth,trim={5cm 1cm 5cm 1cm},clip]{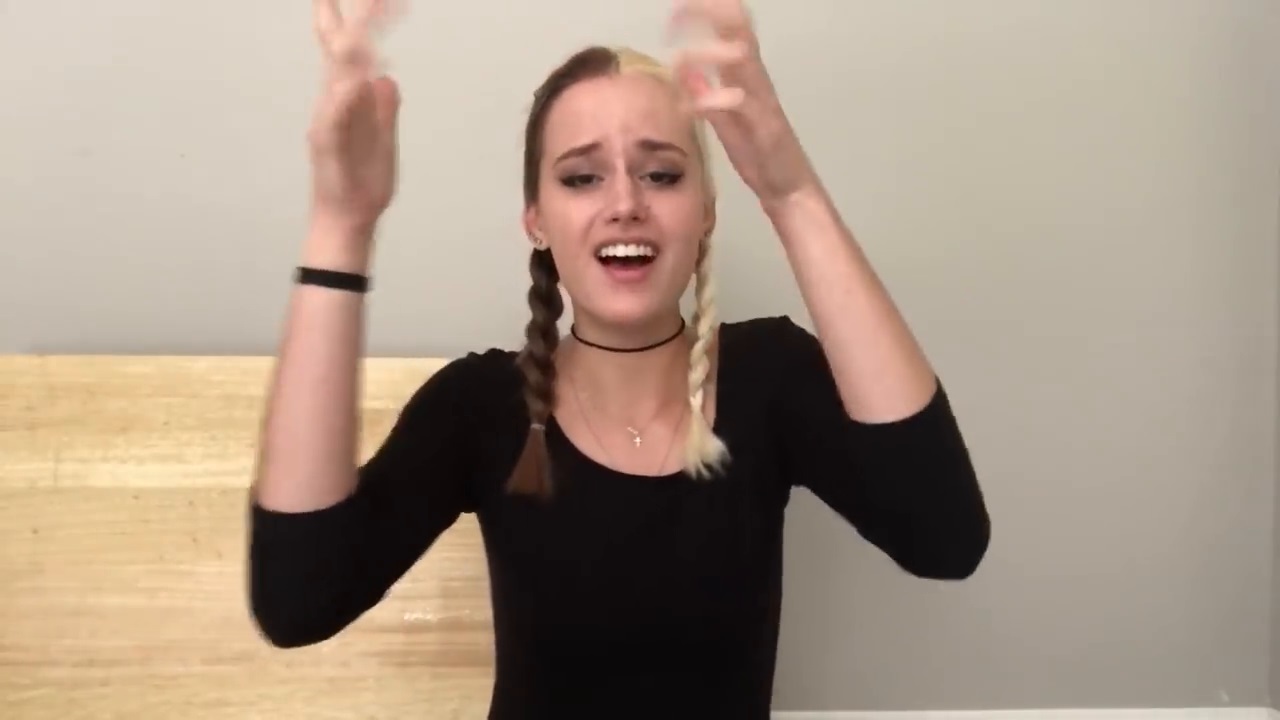}\hspace{-1em}
    \includegraphics[width=0.13\textwidth,trim={5cm 1cm 5cm 1cm},clip]{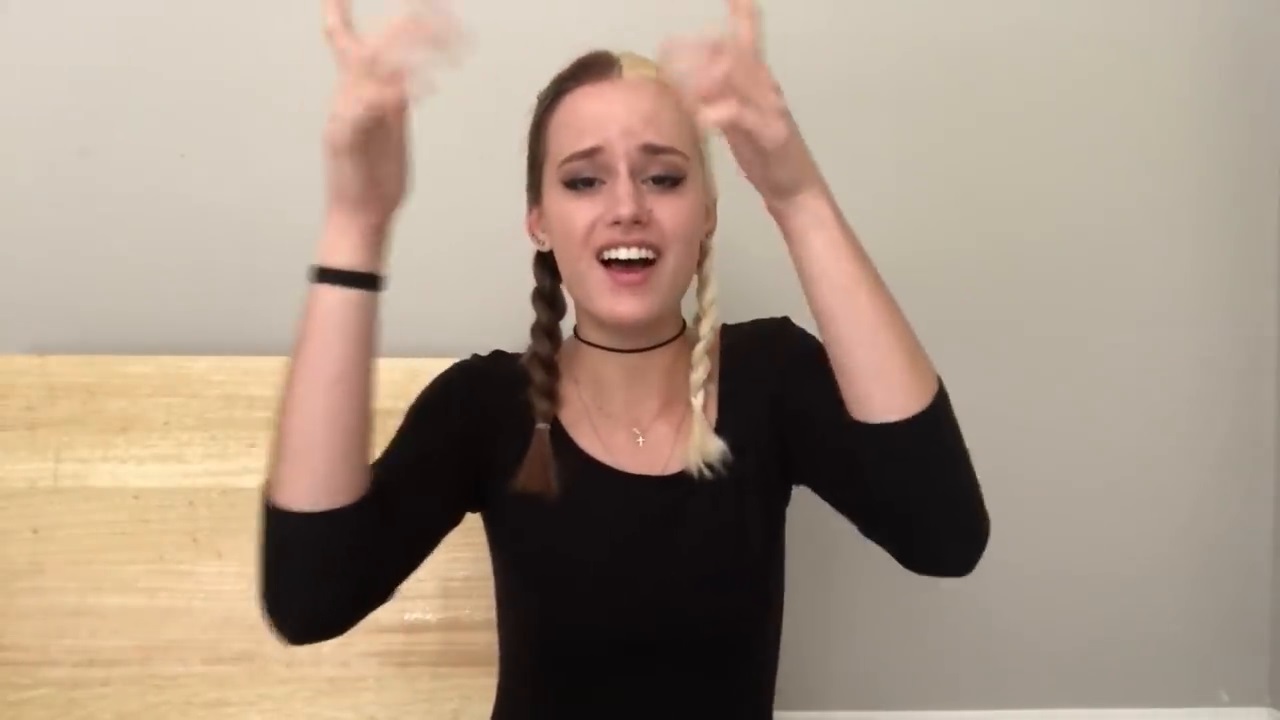}\hspace{-1em}
    \includegraphics[width=0.13\textwidth,trim={5cm 1cm 5cm 1cm},clip]{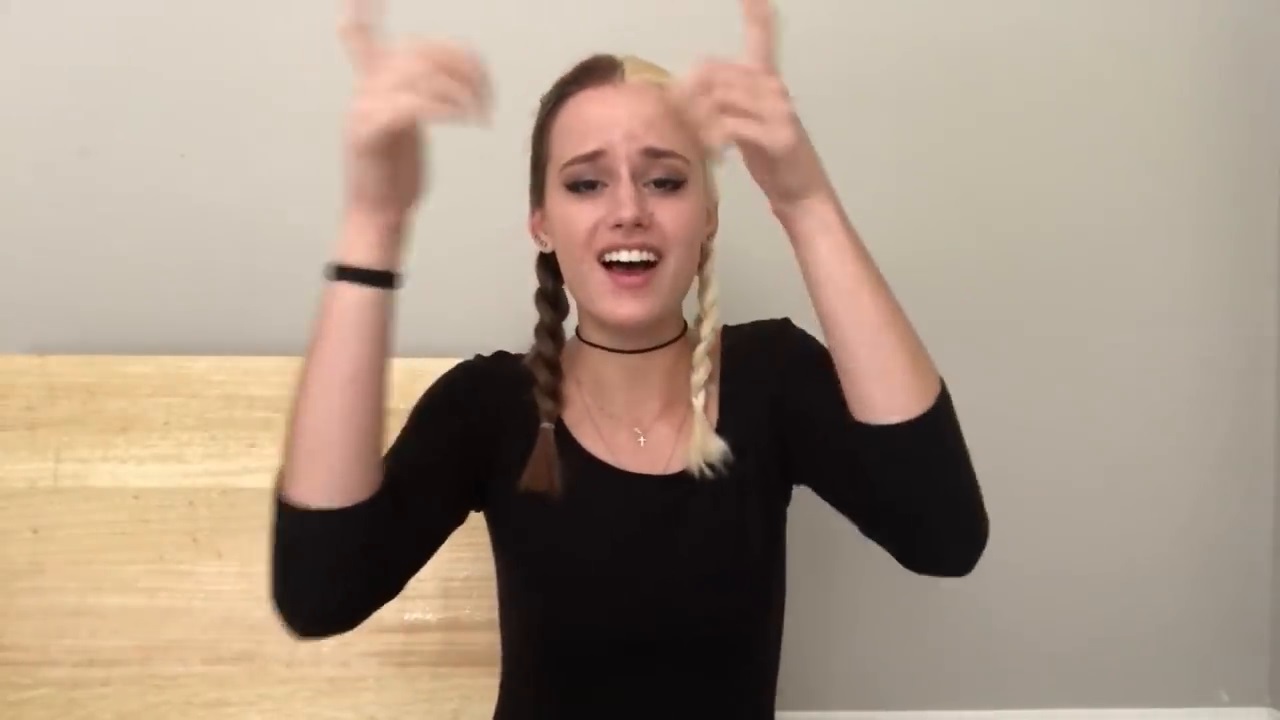}
    
    \shortstack[c]{\vspace{0.7cm}(b)} \includegraphics[width=0.13\textwidth,trim={5cm 1cm 5cm 1cm},clip]{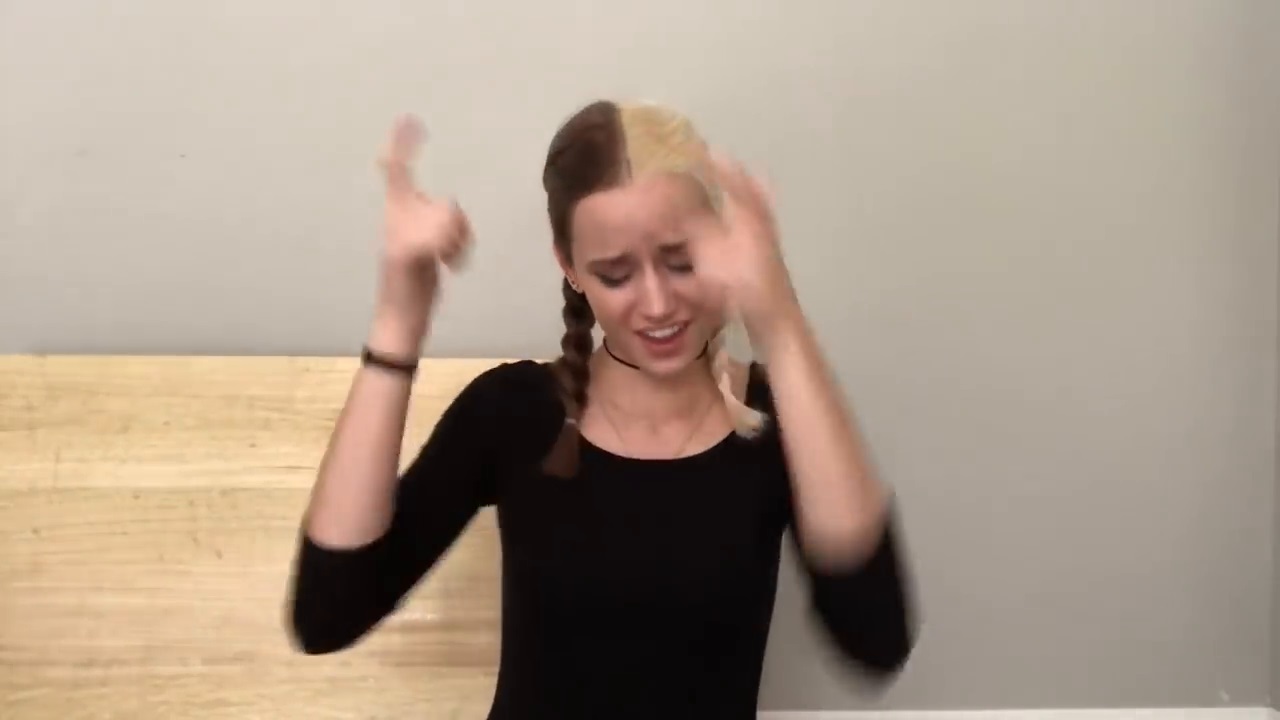}\hspace{-1em}
    \includegraphics[width=0.13\textwidth,trim={5cm 1cm 5cm 1cm},clip]{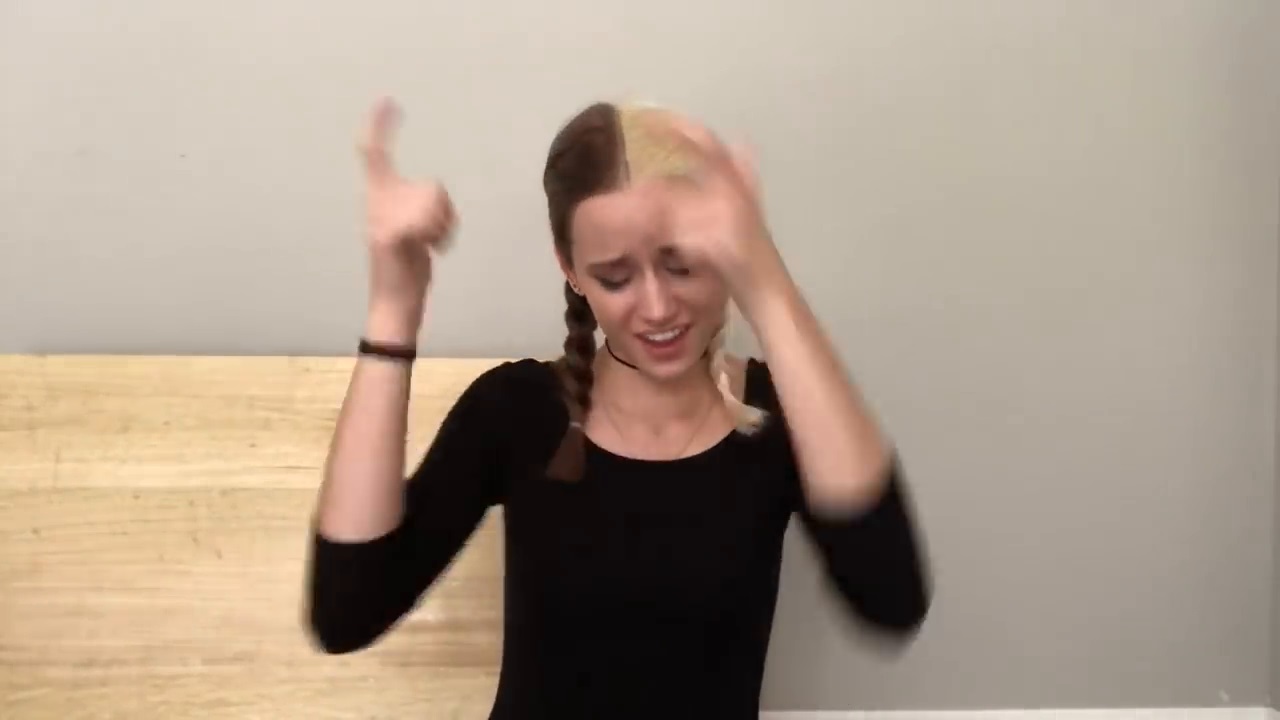}\hspace{-1em}
    \includegraphics[width=0.13\textwidth,trim={5cm 1cm 5cm 1cm},clip]{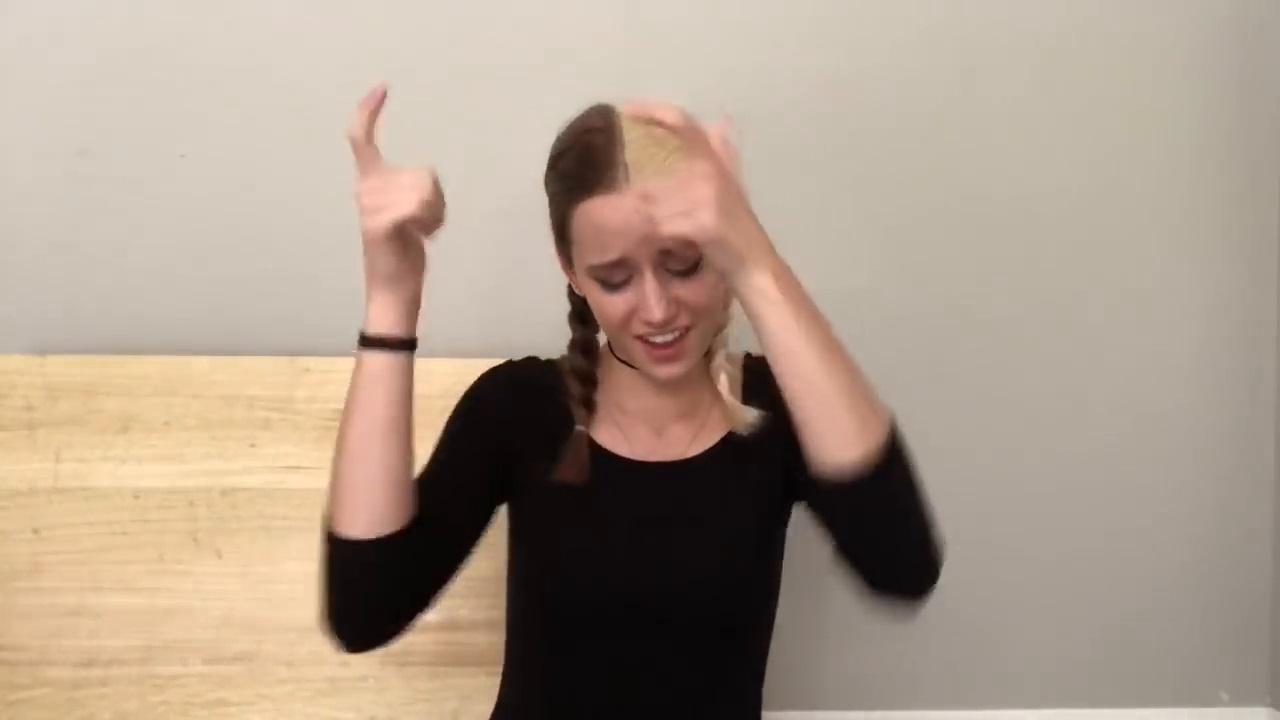}\hspace{-1em}
    \includegraphics[width=0.13\textwidth,trim={5cm 1cm 5cm 1cm},clip]{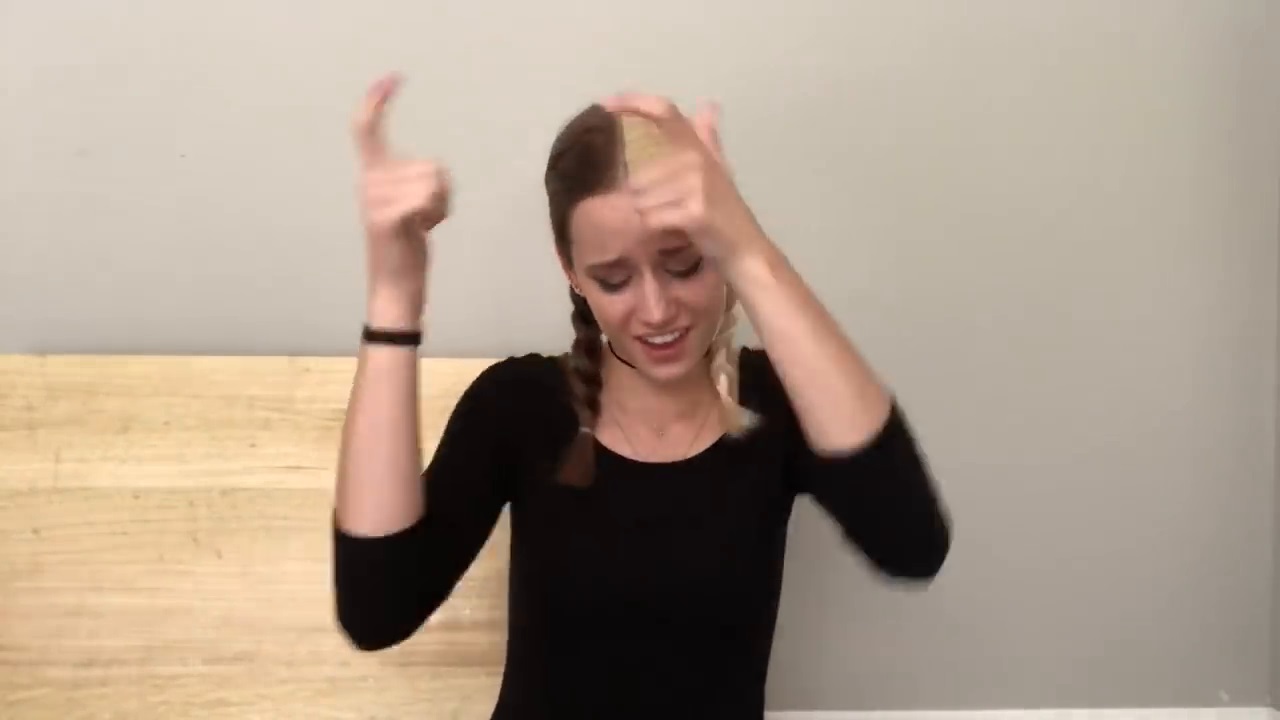}\hspace{-1em}
    \includegraphics[width=0.13\textwidth,trim={5cm 1cm 5cm 1cm},clip]{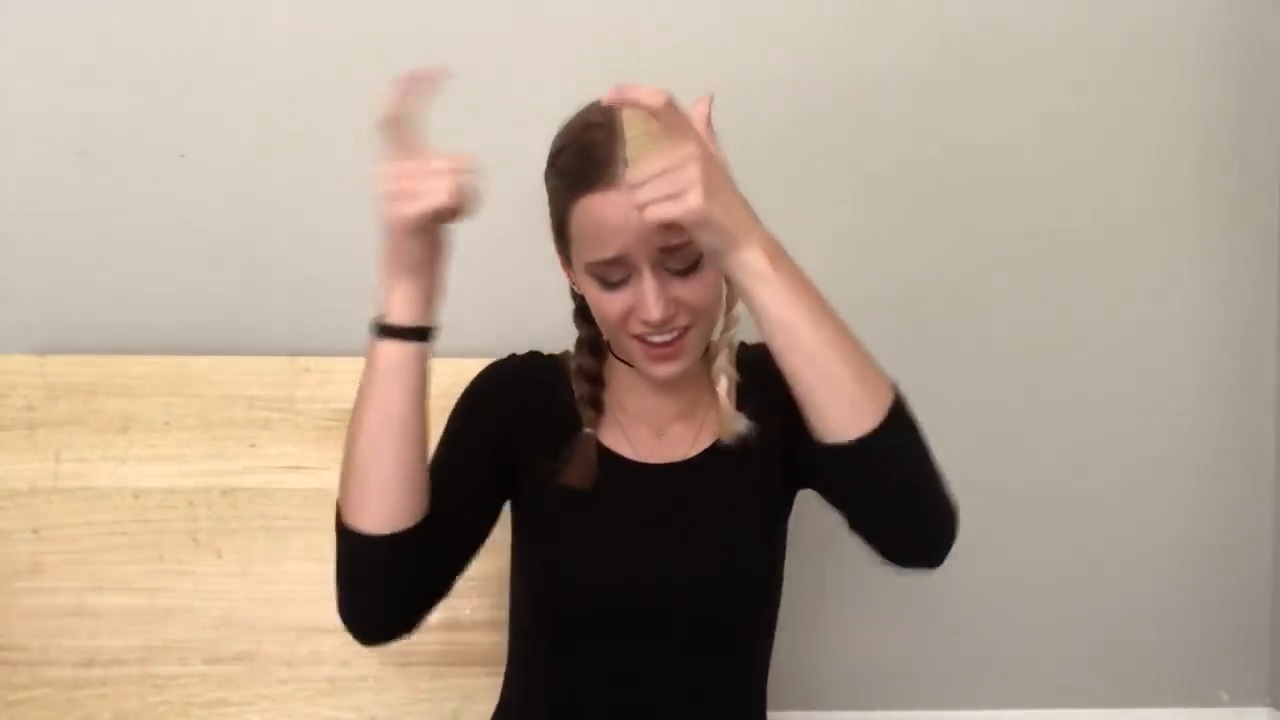}\hspace{-1em}
    \includegraphics[width=0.13\textwidth,trim={5cm 1cm 5cm 1cm},clip]{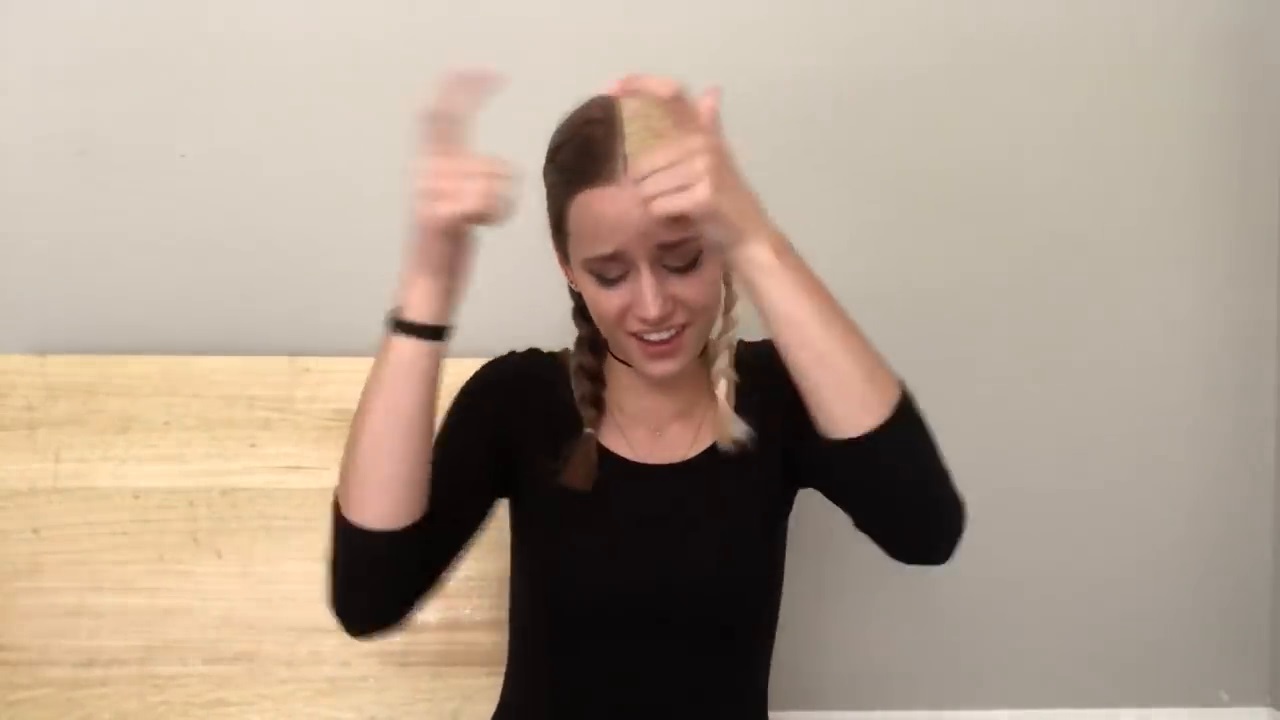}\hspace{-1em}
    \includegraphics[width=0.13\textwidth,trim={5cm 1cm 5cm 1cm},clip]{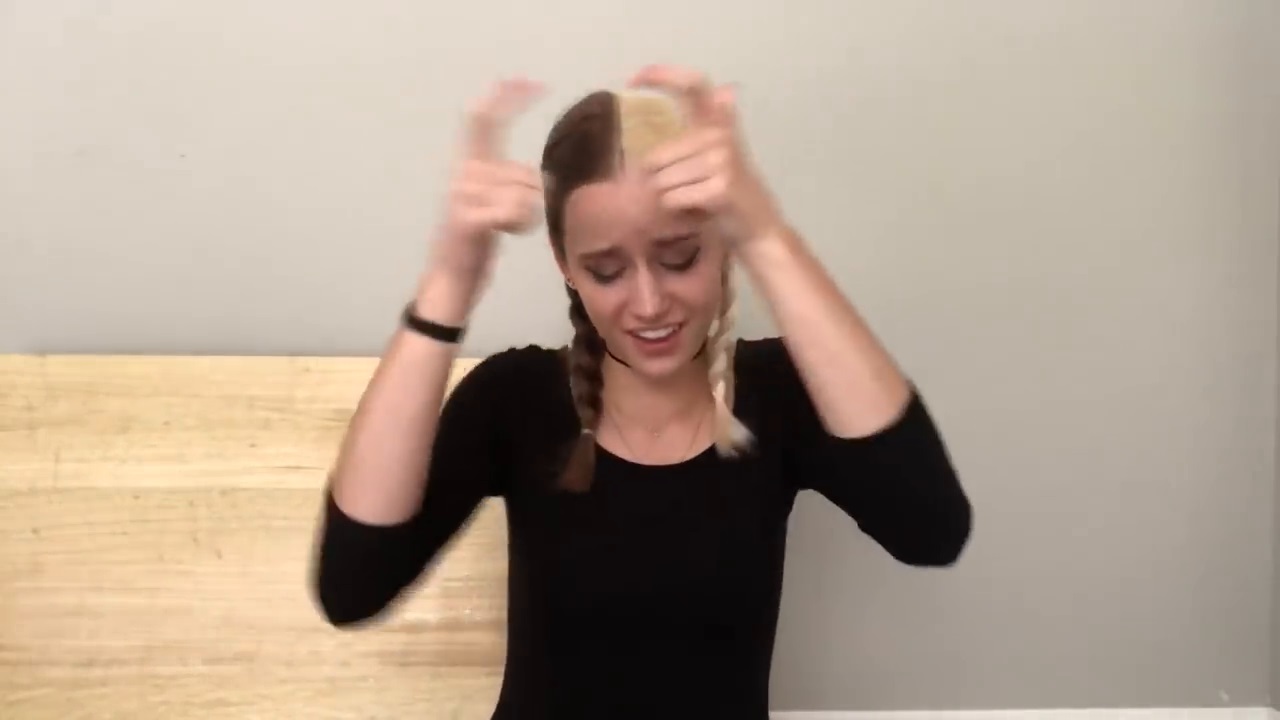}
    \caption{Two similar phonemes (Frames 4736-4742 for the phoneme (a) \& 7126-7132 for the phoneme (b)) from cluster B (violet). Phonemes (a) and (b) being a part of the `Halo, halo' lyrics.}
    \label{fig:visual2}
\end{figure}

From the Figures \ref{fig:visual1} and \ref{fig:visual2} it can be seen that the phonemes are visually similar where every phoneme is described with the location and orientation phonological parameters. It is worth noting that the meaning can or cannot be the same, but the phonological parameters are similar having at least 50\% phoneme similarity.

\subsection{Matching Consecutive Sequences of Phonemes}
Single video is chosen to search for similar consecutive phonemes in different parts of the video. The lyrics in the background music in the video are used as the ground truth for verifying the results. Phoneme similarity threshold is chosen to be 50\%.


\begin{figure}[H]
    \centering
    \shortstack[c]{\vspace{0.7cm}(a)} \includegraphics[width=0.13\textwidth,trim={5cm 1cm 5cm 1cm},clip]{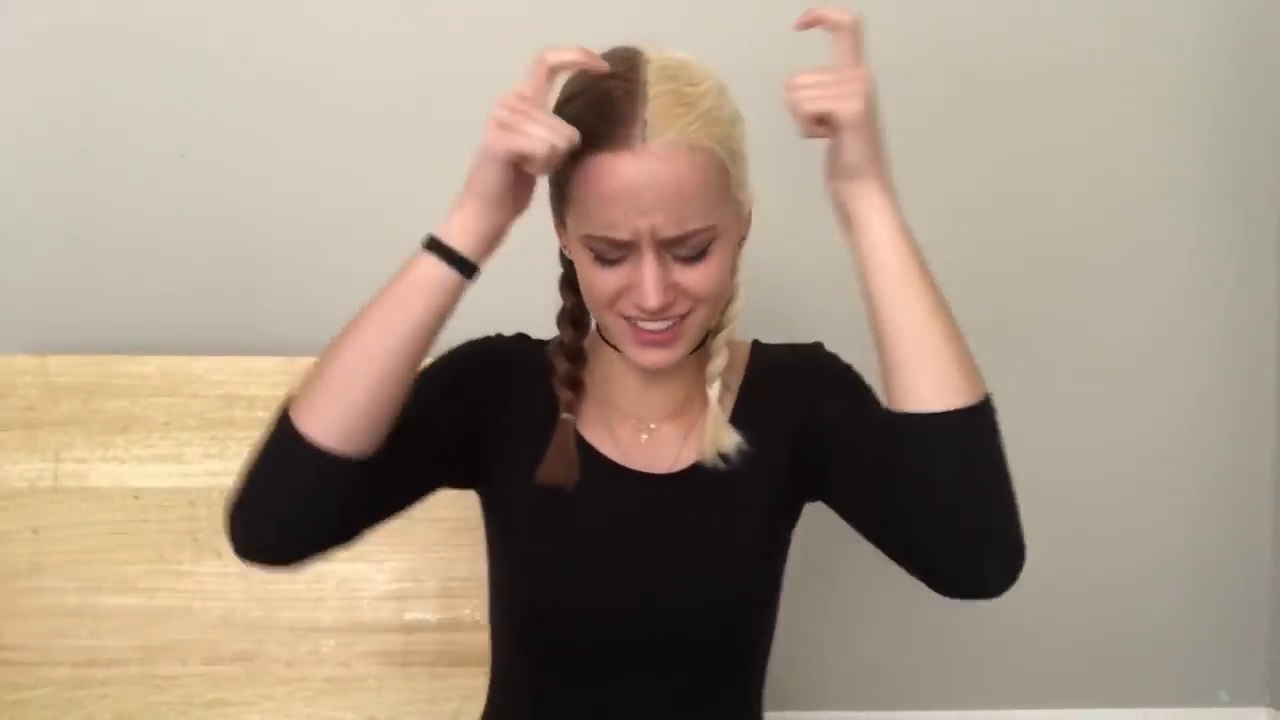}\hspace{-1em}
    \includegraphics[width=0.13\textwidth,trim={5cm 1cm 5cm 1cm},clip]{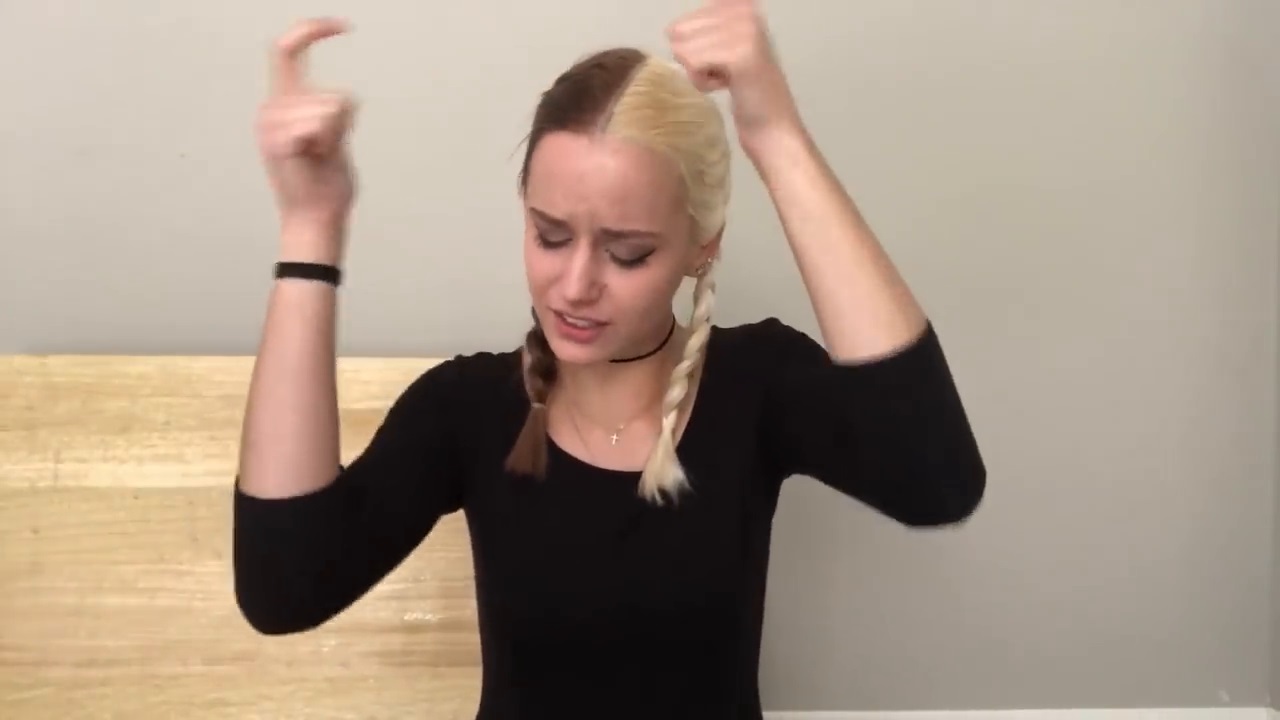}\hspace{-1em}
    \includegraphics[width=0.13\textwidth,trim={5cm 1cm 5cm 1cm},clip]{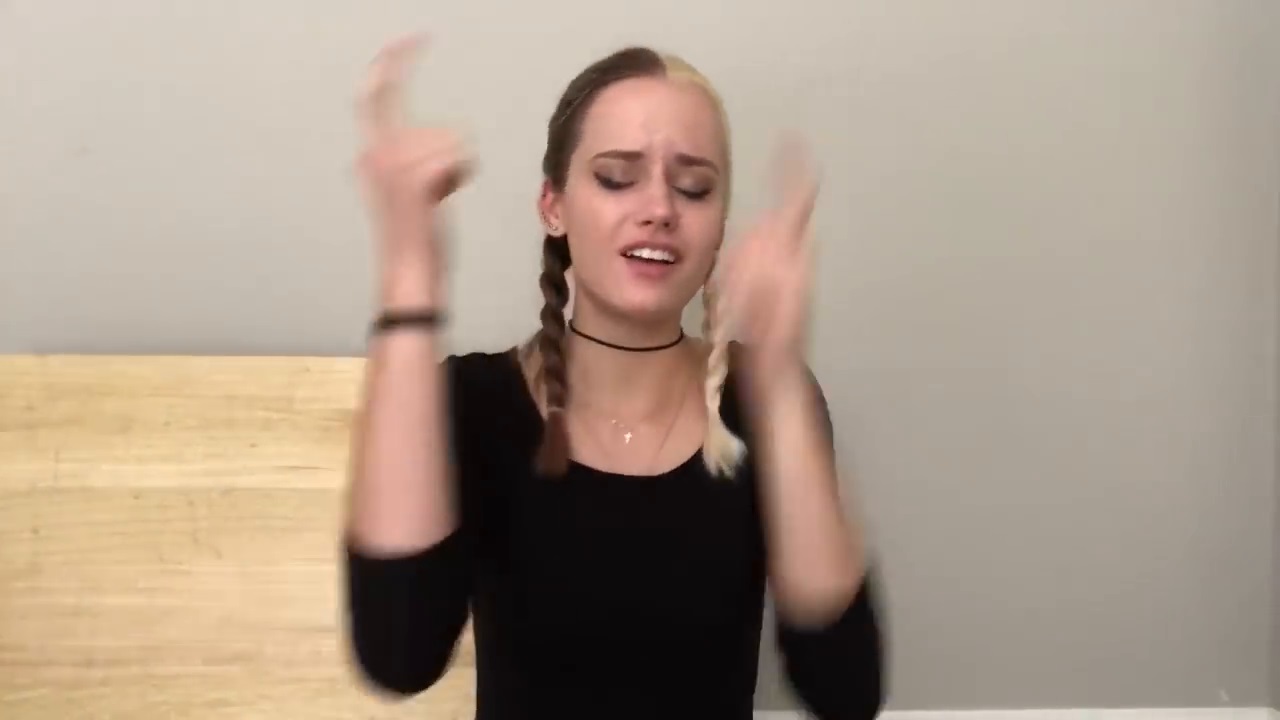}\hspace{-1em}
    \includegraphics[width=0.13\textwidth,trim={5cm 1cm 5cm 1cm},clip]{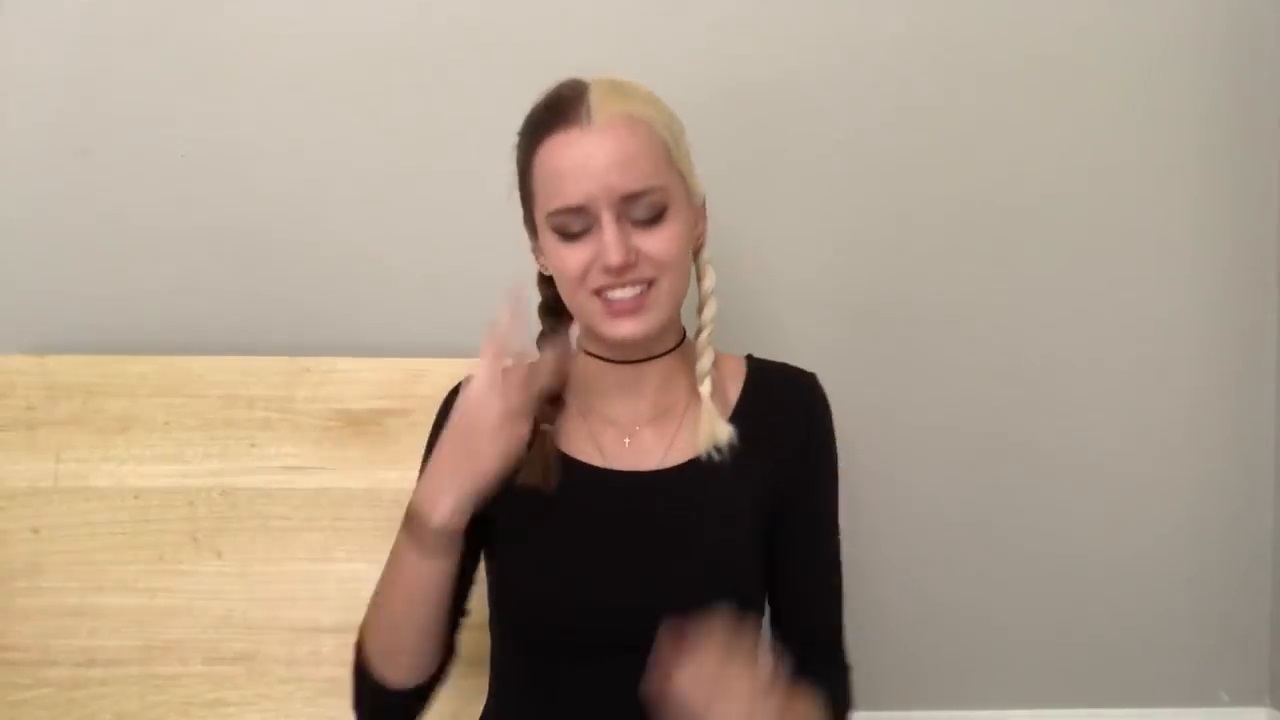}\hspace{-1em}
    \includegraphics[width=0.13\textwidth,trim={5cm 1cm 5cm 1cm},clip]{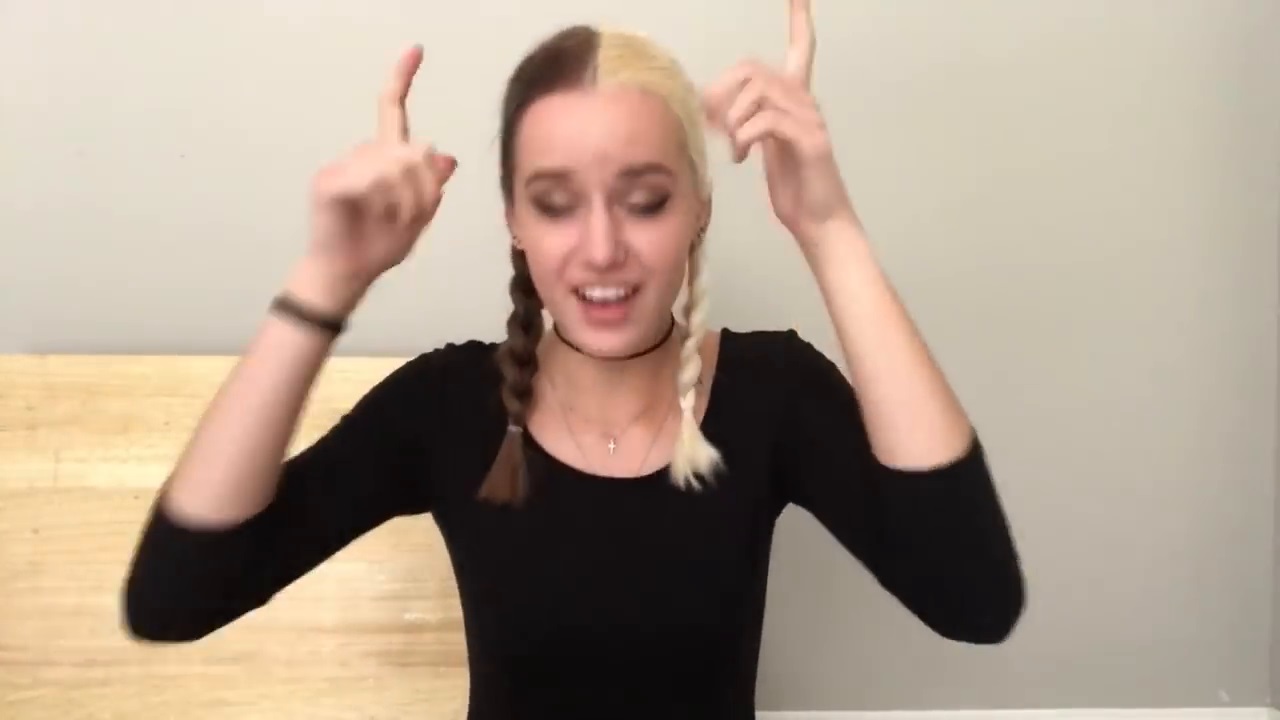}\hspace{-1em}
    \includegraphics[width=0.13\textwidth,trim={5cm 1cm 5cm 1cm},clip]{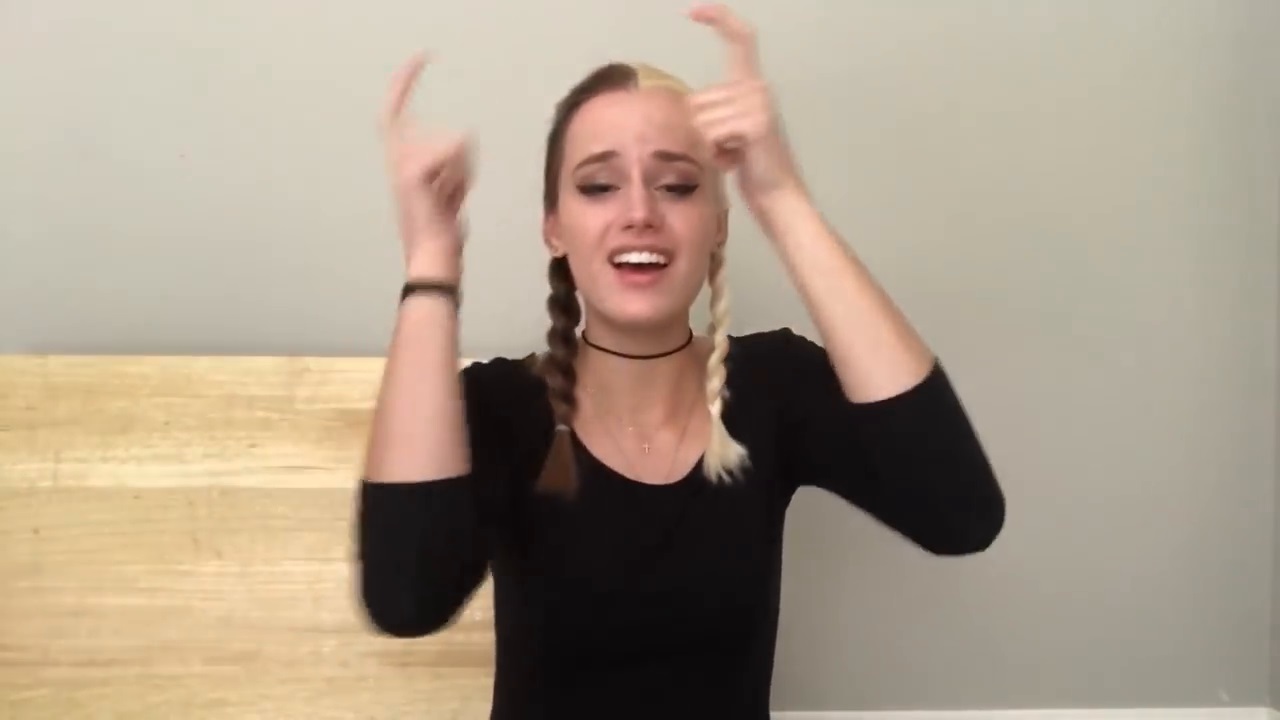}\hspace{-1em}
    \includegraphics[width=0.13\textwidth,trim={5cm 1cm 5cm 1cm},clip]{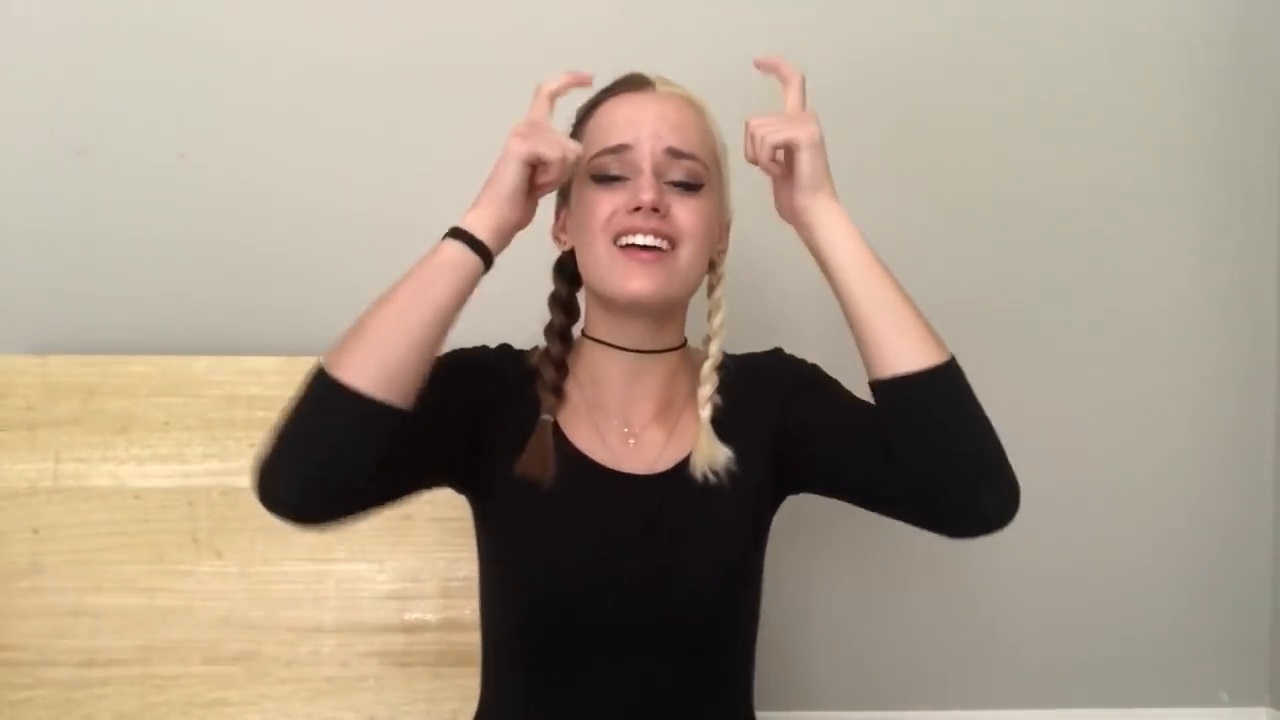}\hspace{-1em}
    
    \shortstack[c]{\vspace{0.7cm}(b)} \includegraphics[width=0.13\textwidth,trim={5cm 1cm 5cm 1cm},clip]{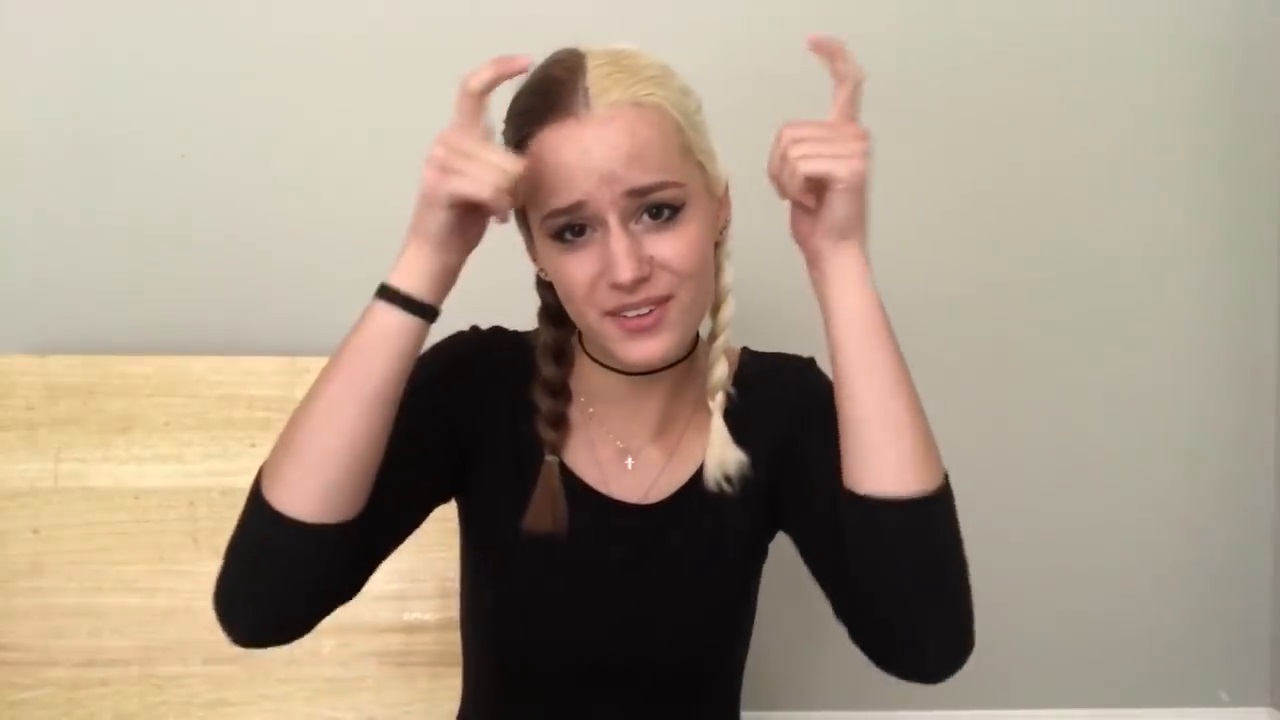}\hspace{-1em}
    \includegraphics[width=0.13\textwidth,trim={5cm 1cm 5cm 1cm},clip]{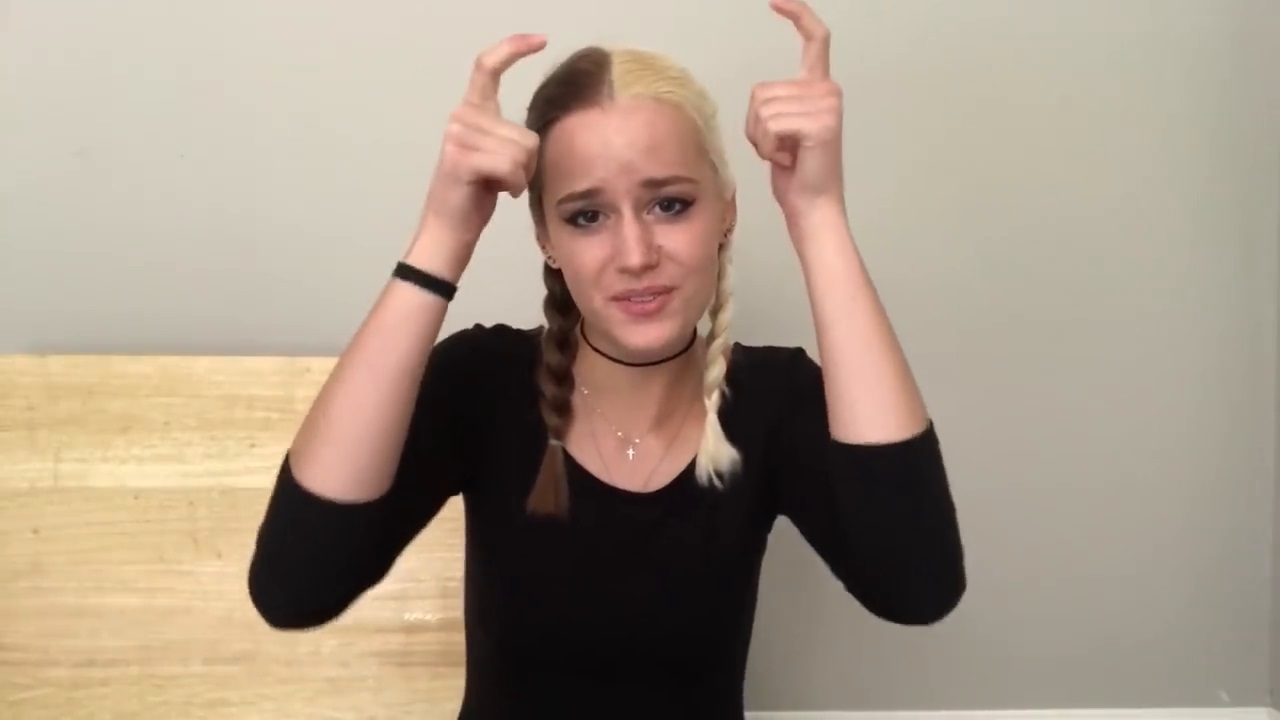}\hspace{-1em}
    \includegraphics[width=0.13\textwidth,trim={5cm 1cm 5cm 1cm},clip]{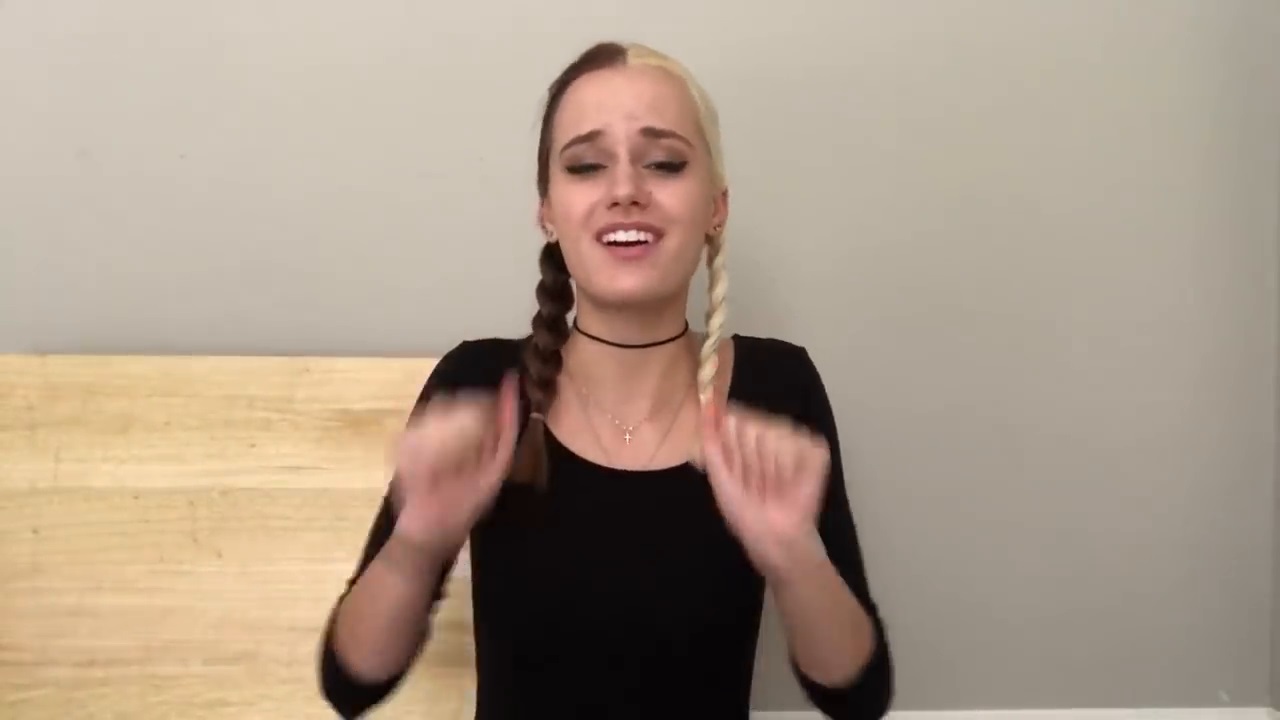}\hspace{-1em}
    \includegraphics[width=0.13\textwidth,trim={5cm 1cm 5cm 1cm},clip]{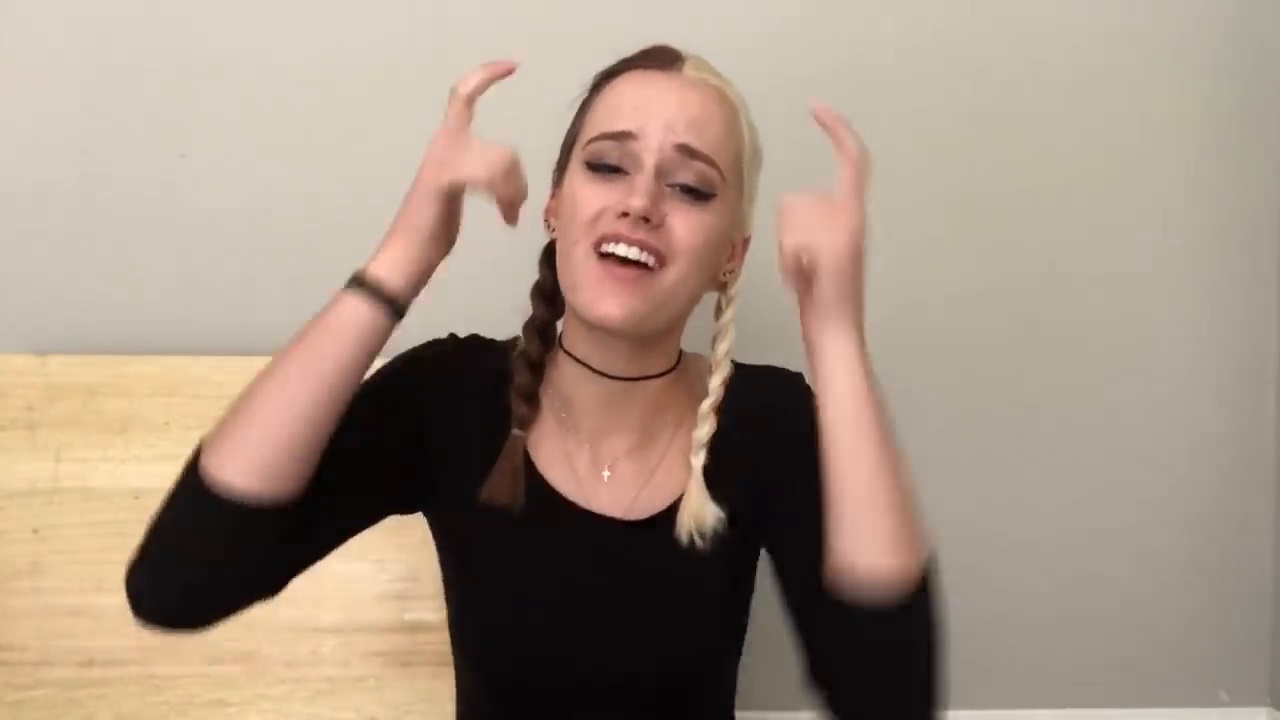}\hspace{-1em}
    \includegraphics[width=0.13\textwidth,trim={5cm 1cm 5cm 1cm},clip]{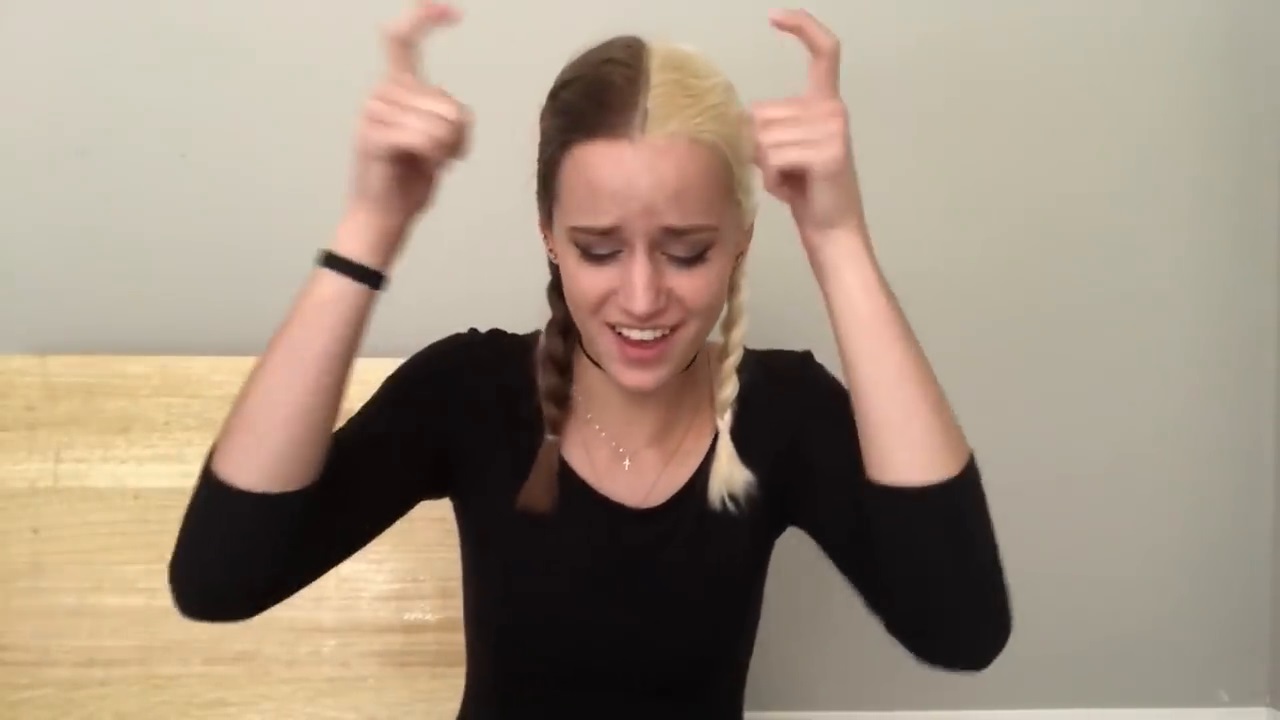}\hspace{-1em}
    \includegraphics[width=0.13\textwidth,trim={5cm 1cm 5cm 1cm},clip]{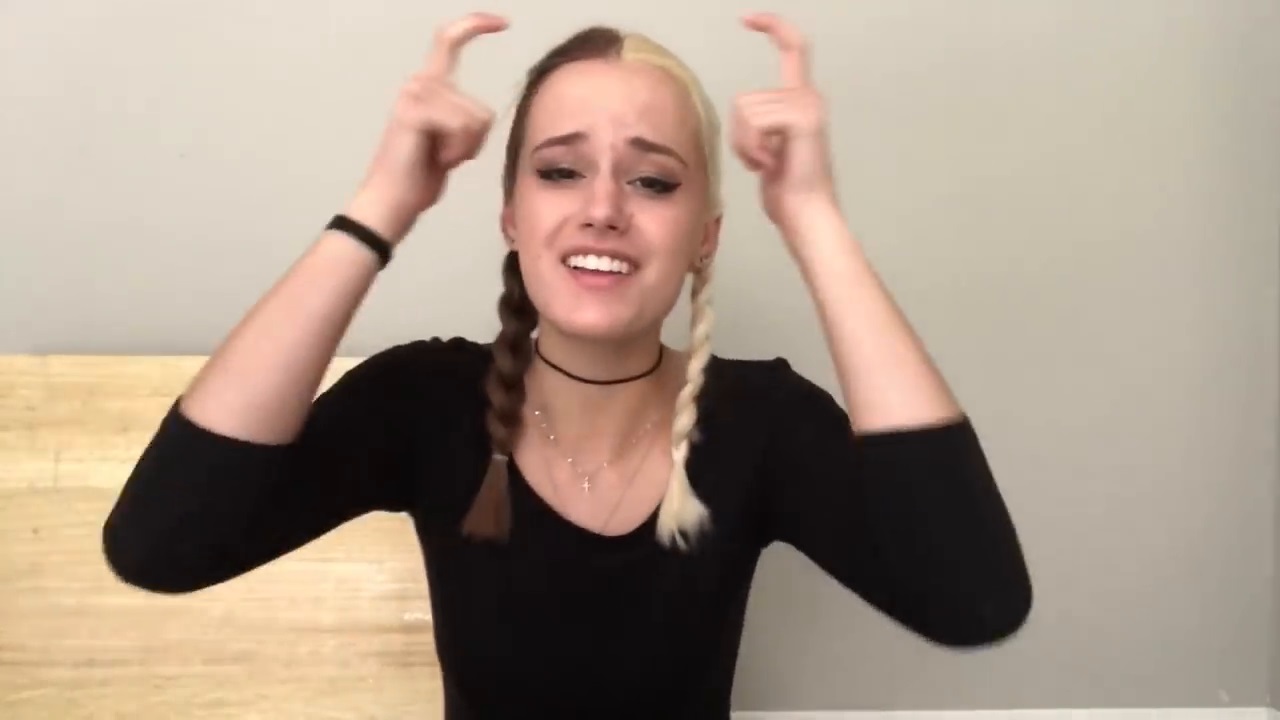}\hspace{-1em}
    \includegraphics[width=0.13\textwidth,trim={5cm 1cm 5cm 1cm},clip]{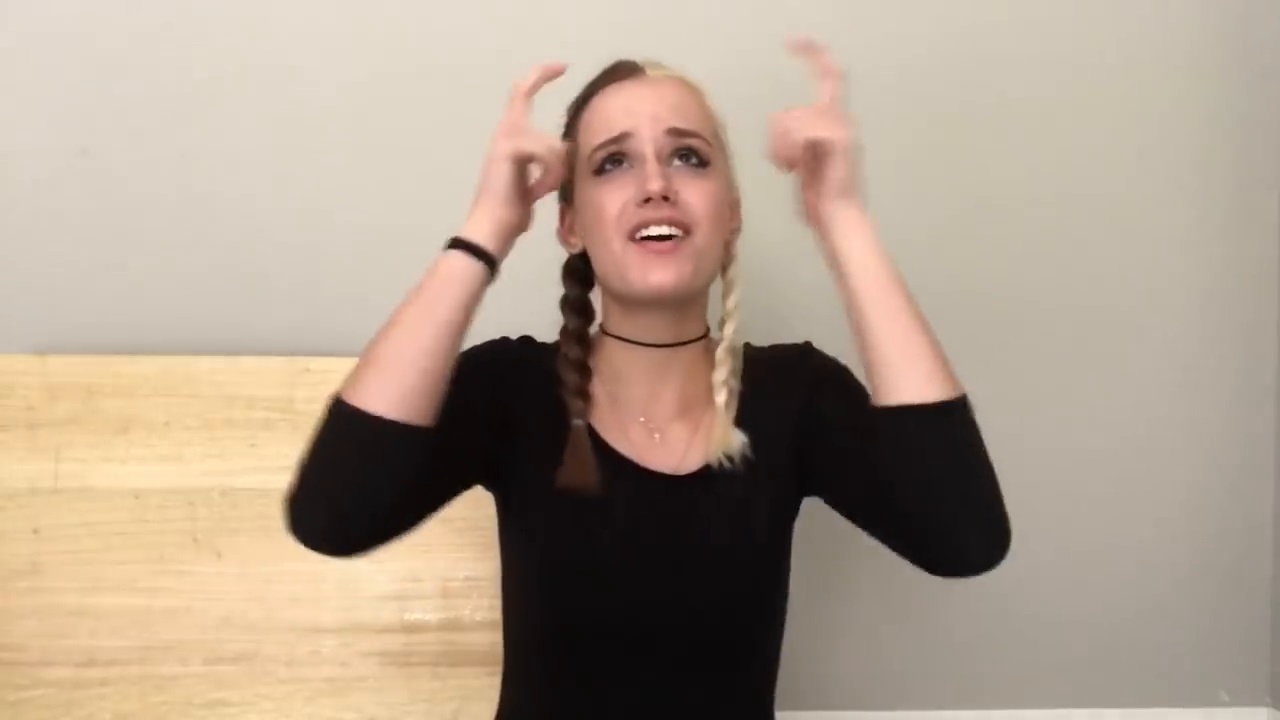}\hspace{-1em}
    
    \caption{Two similar combinations of consecutive phonemes (Frames 5012-5207 for consecutive phonemes (a) \& 2451-2839 for consecutive phonemes (b)) corresponding to the same lyrics:}
    \label{fig:visual4}
\end{figure}

\begin{table}[H]
\vspace{-2em}
    \centering
    \begin{tabular}{l}
        \qquad\qquad\qquad\qquad\qquad\qquad\qquad  ``I can feel your Halo, Halo, Halo. I can see your Halo, Halo, Halo \\
        \qquad\qquad\qquad\qquad\qquad\qquad\qquad  \; I can feel your Halo, Halo, Halo. I can see your Halo, Halo, Halo''
    \end{tabular}
    \caption*{}
\end{table}


As a result it can be seen that the same phoneme similarity approach described in Section 4.4 can be used to find similar consecutive phonemes in the same video as shown in Figure \ref{fig:visual4}, where consecutive phonemes in (a) and consecutive phonemes in (b) correspond to the same lyrics from the song.

\section{Conclusion}
This work explored phoneme clustering for continuous signing videos obtained online using only location and orientation phonological parameters for phoneme comparison. An iterative grouping method and the DBSCAN methods were compared, showing that the DBSCAN is more unpredictable for the same signed content when the phoneme similarity conditions (neighbourhood) are relaxed. Nevertheless, DBSCAN has a more even distribution of identified clusters and the average cluster size when similarity conditions are relaxed as compared to the iterative grouping method. This work has shown that it is possible to use a linguistically viable distance metric for general purpose clustering algorithms to work on sign language data. To improve the clustering approach, it would be beneficial to include other phonological parameters, such as handshape and even non-manual features. In the future all the obtained interpretation videos for the same song will be clustered in order to compare phonemes from the same clusters across the different signers. 

The benefit of using clustering on music videos is that it is possible to identify repeating patterns in videos, which correspond to verses in music. This knowledge can be used to generate datasets for the computational sign language models as more and more song interpretations in sign languages performed by different signers become available online. Moreover, such phoneme clustering approaches could be used during the resource annotation for linguistic purposes. Similar phonemes, signs, and even phrases could be identified in continuous signing videos prior to watching such videos.

\bibliographystyle{coling}
\bibliography{coling2020}

\end{document}